\PassOptionsToPackage{table}{xcolor} 
\documentclass[11pt]{article}
\usepackage[final]{acl}   
\usepackage{times}
\usepackage{latexsym}
\usepackage[T1]{fontenc}
\usepackage[utf8]{inputenc}
\usepackage{microtype}
\IfFileExists{inconsolata.sty}{\usepackage{inconsolata}}{}
\usepackage{xcolor}
\usepackage{graphicx}
\usepackage{tikz}
\usepackage{booktabs}
\usepackage{adjustbox} 
\usepackage{multirow}
\usepackage{placeins}
\usepackage{amssymb}
\IfFileExists{xurl.sty}{\usepackage{xurl}}{}
\usepackage{fontawesome5}
\graphicspath{{figures/}}

\newcommand{\keyrow}{\rowcolor{gray!18}} 

\newif\ifanon \anonfalse   
\newcommand{\named}[2]{\ifanon #2\else #1\fi}
\newcommand{\ci}[2]{[$#1$,\,$#2$]}
\newcommand\blfootnote[1]{\begingroup\renewcommand\thefootnote{}\footnote{#1}\addtocounter{footnote}{-1}\endgroup}

\title{Lost in Refusal: How Far Can Guard-Measured Safety Drift\\ Be Trusted After Benign Multilingual Fine-Tuning?}

\author{\named{%
  \textbf{Aishwarya Nikkumbh}\textsuperscript{$\ast$},
  \textbf{Mingyu Kang}\textsuperscript{$\ast$},
  \textbf{Hannah Pham} \\
  Purdue University \\[2pt]
  \small\faEnvelope\hspace{0.4em}\texttt{anikkumb@purdue.edu, kang719@purdue.edu, pham215@purdue.edu} \\
  \small\faGithub\hspace{0.4em}\href{https://github.com/lost-in-refusal/lexsi-multilingual-safety}{Code}
  \hspace{1.2em}\faDatabase\hspace{0.4em}\href{https://huggingface.co/collections/lost-in-refusal/lost-in-refusal-6b07c29a5b461cd1af8df254}{Data and models}}{Anonymous submission}}

\begin{document}
\raggedbottom
\maketitle
\named{\blfootnote{$^\ast$Equal contribution.}}{}

\begin{abstract}
Teams that fine-tune open multilingual models on harmless data usually re-check safety with an automatic classifier, or \emph{guard}. We find that such a check detects real damage, including in languages the team never trained on, but overstates it. Fine-tuning Tiny Aya Global (3.35B) on harmless English, Korean and Chinese instructions, under a preregistered three-seed design, raises the share of Korean answers to MultiJail attack prompts that Qwen3Guard flags from 5.2\% to 17.9\%, with no detected change in English, and the rise persists when no Korean rows are used in training. Three native Korean readers outside the team, reading with the training condition hidden, each confirm the rise, but at about 40\% to 75\% of the guard's size. The guard also flags 5.5 points more of the fine-tuned model's Korean answers to harmless prompts, suggesting it reacts partly to how the fine-tuned model answers rather than to harm alone. Mixing five copies of a few dozen of the base model's own refusals into training lowers the Korean rise more than gradient surgery (SafeGrad), at the cost of more refusals of harmless requests. We release all answers, guard verdicts, adapters and the preregistration with its failed checks.
\end{abstract}


\section{Introduction}

Small open multilingual models are built to be adapted. Tiny Aya Global \citep{salamanca2026tinyaya}, a 3.35B-parameter model, is a typical case: a business fine-tunes it on its own harmless data, say for a support assistant in English, Korean and Chinese, and serves it in all three. Safety is then checked again, usually with an automatic safety classifier (a \emph{guard}) and often only in English or in the training languages.

Fine-tuning on harmful, and even on benign, data can undo safety training \citep{qi2024finetuning,he2024safedata}, and models are less safe in lower-resource languages \citep{deng2024multijail,shen2024languagebarrier}; a fine-tuning attack in one language carries over to others \citep{poppi2025fragility}.

\citet{hawkins2026heterogeneous} show that a benign fine-tune changes safety unevenly across fine-tuning and evaluation languages, scoring non-English answers by translating them back for an English judge. How much of such a guard-measured change is harm, rather than a change in how the guard reads the fine-tuned model's answers, is rarely checked against people. We test this on one model family and one guard, under a preregistered design, with native speakers reading the answers blind, and test two repairs. We find that:
\begin{enumerate}\setlength\itemsep{0pt}\setlength\parsep{0pt}
\item The guard-flagged rate rises unevenly by language: in Korean from 5.2\% to 17.9\%, with no detected change in English, and Korean training rows are not needed for the rise (\S\ref{sec:drift}, \S\ref{sec:language}).
\item People confirm the rise, but the guard overstates it. Three of four blinded readers detect it on Tiny Aya, at about 40\% to 75\% of the guard's size, the guard also flags more answers to harmless prompts, and on a second model family two of four readers detect it (\S\ref{sec:measure}, \S\ref{sec:granite}).
\item The rise also appears on a fine-tune that improves the task it is trained for, so it is not only general damage, though the general task score falls there too (\S\ref{sec:damage}).
\item Mixing five copies of a few dozen of the base model's own refusals into the training data \citep{bianchi2024safetytuned} lowers the Korean rise more than SafeGrad \citep{yi2025safegrad}, at the cost of more refusals of harmless requests (\S\ref{sec:repair}).
\end{enumerate}
We release the answers, guard verdicts and adapters, the preregistration with its failed checks, the final model,\named{\footnote{The final model, the instruction fine-tune with the refusals mixed in: \url{https://huggingface.co/lost-in-refusal/tiny-aya-global-bactrian-lora-mix5}.}}{} and 26 issues filed against the tools we used (Appendices~\ref{app:feedback}, \ref{app:repro}).

\section{Related Work}
\label{sec:related}

\paragraph{Fine-tuning erodes safety.} Fine-tuning on a few harmful examples, and even on benign data, can undo safety training \citep{qi2024finetuning,yang2023shadow,lermen2023lora,zhan2024removing,wei2024brittleness}; \citet{he2024safedata} identify which benign data does the most harm.

\paragraph{Multilingual safety.} Models are less safe in lower-resource languages \citep{deng2024multijail,yong2023lowresource,wang2024alllanguages,shen2024languagebarrier}, and a fine-tuning attack in one language carries over to others \citep{poppi2025fragility}; benign fine-tuning also shifts safety unevenly across fine-tuning and evaluation languages \citep{hawkins2026heterogeneous}. We ask how far a guard's reading of that shift can be trusted, language by language.

\paragraph{Guards and judges.} Safety evaluations increasingly score answers with a guard model such as Qwen3Guard \citep{qwen2025guard}, PolyGuard \citep{kumar2025polyguard}, Granite Guardian \citep{padhi2024graniteguardian} or WildGuard \citep{han2024wildguard}, or with a model that follows a written policy, such as gpt-oss-safeguard \citep{openai2025safeguard}. We check one guard against human labels, three other judges and blinded human readers, and test whether it reacts to fine-tuned answers to harmless prompts.

\paragraph{Repairs.} Proposed defences mix safety examples into the fine-tuning data \citep{bianchi2024safetytuned}, harden alignment or project the update \citep{huang2024vaccine,hsu2024safelora}, or apply gradient surgery against an alignment loss \citep{yi2025safegrad}. We compare mixing with SafeGrad.

\section{Setup}
\label{sec:setup}

\paragraph{Models and data.} Tiny Aya Global (3.35B, revision \texttt{af89d219}; CC-BY-NC); Tiny Aya Earth (\texttt{0624fafb}) is a regional variant of it. We train on 1,800 harmless Bactrian-X rows \citep{li2023bactrianx}: 600 instructions, each in English, Korean and Chinese, cleaned with CuratorKIT \citep{lexsi2026curatorkit} and screened by our guard, which removed 26 flagged or refusing items. A second dataset is 519 human-written or human-edited Aya rows \citep{singh2024aya}, 173 per language. Swahili is never in training.

\paragraph{Arms.} \emph{Base} (untuned); \emph{plain} LoRA \citep{hu2022lora}; \emph{SafeGrad} LoRA through SafeTune \citep{lexsi2026safetune}: gradient surgery against an alignment loss on 47 of the base model's refusals to held-out MultiJail prompts (18 English, 13 Korean, 16 Chinese). Same data, order, rank 16 and 3 epochs in every arm; seeds 42, 1, 2 (Appendix~\ref{app:details}).

\paragraph{Evaluation.} 291 MultiJail \citep{deng2024multijail} \emph{prompt families}, a family being one prompt in English, Korean, Chinese and Swahili. The prompts are direct harmful requests; nothing adapts to the model. 250 safe XSTest prompts per language (249 in Swahili) measure over-refusal \citep{rottger2024xstest}, and 300 held-out Bactrian-X prompts the task. Decoding is greedy, capped at 256 new tokens. A second benchmark is the 200 unsafe XSTest prompts, machine-translated with NLLB-200 \citep{nllb2022}.

\paragraph{Metrics.} Every safety metric is a label of one guard, Qwen3Guard-Gen-4B \citep{qwen2025guard}, which gives each answer a safety label and a refusal label. The \emph{guard-flagged rate} is the share of answers to attack prompts labelled Unsafe or Controversial (what MultiJail calls attack success); a flagged answer is not shown to be usable harm. The same share on harmless prompts is the \emph{harmless-prompt flag rate}. The \emph{refusal rate} is the share of answers to attack prompts with the refusal label, and \emph{over-refusal} the same share on safe XSTest prompts. The \emph{task score}, 1 to 7, is judged by Qwen3-14B \citep{qwen2025qwen3}. Changes are in points on the same prompts, and each block of runs is compared with its own base run (Appendix~\ref{app:details}).

\paragraph{Statistics.} Per language and seed we use exact McNemar tests. Pooled effects use a bootstrap over prompts (5,000 draws, 95\% intervals), each prompt's value averaged over the seeds first. The unit resampled is the prompt (on MultiJail the 291 families, a contrast between two languages being taken inside a family). A pooled interval is conditional on the seeds we trained: it covers the sampling of prompts, not the spread between seeds. Nothing is corrected for the number of checks. We call a change \emph{detected} when its 95\% interval excludes zero (per seed, when $p<0.05$) and \emph{not detected} otherwise; not detected does not mean absent.

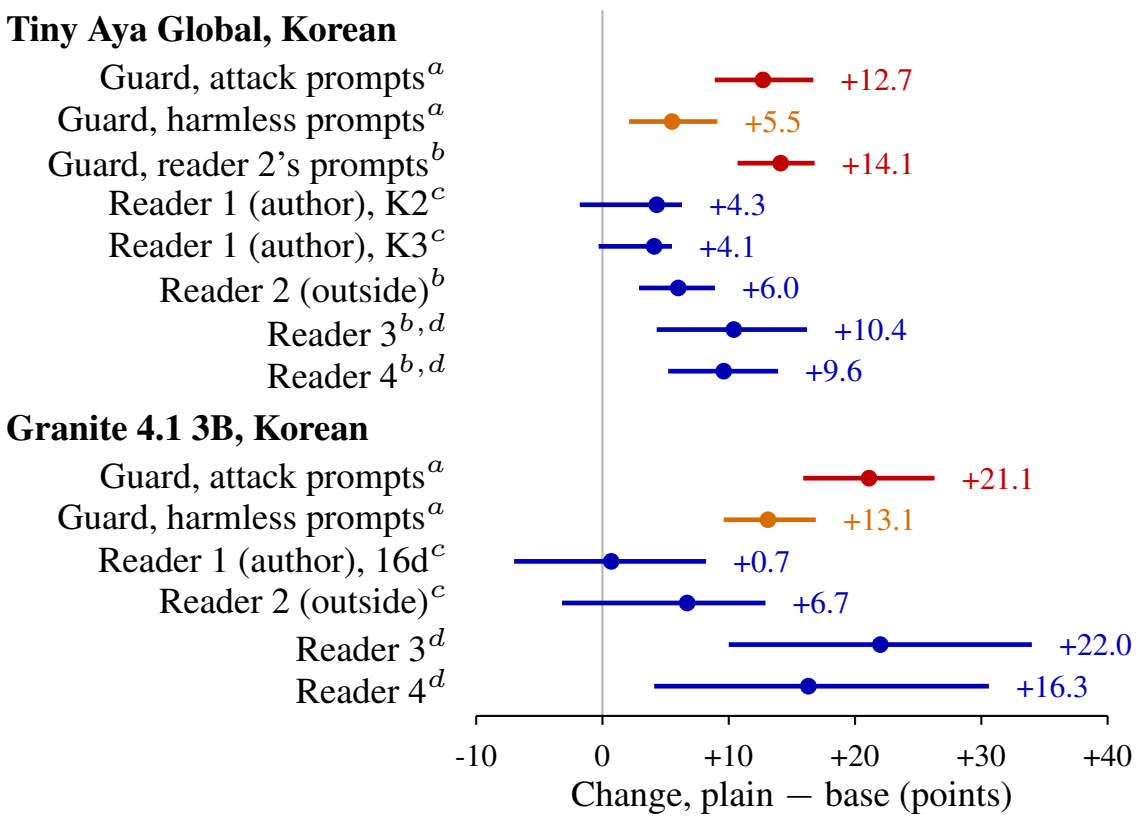
\begin{figure}[t]
\centering
\begin{adjustbox}{max width=\columnwidth}
\begin{tikzpicture}
\draw[gray!60] (0.000,0.15) -- (0.000,-4.60);
\draw (-0.850,-4.60) -- (3.400,-4.60);
\draw (-0.850,-4.60) -- (-0.850,-4.66) node[below,font=\tiny] {-10};
\draw (0.000,-4.60) -- (0.000,-4.66) node[below,font=\tiny] {0};
\draw (0.850,-4.60) -- (0.850,-4.66) node[below,font=\tiny] {+10};
\draw (1.700,-4.60) -- (1.700,-4.66) node[below,font=\tiny] {+20};
\draw (2.550,-4.60) -- (2.550,-4.66) node[below,font=\tiny] {+30};
\draw (3.400,-4.60) -- (3.400,-4.66) node[below,font=\tiny] {+40};
\node[font=\scriptsize] at (1.275,-5.15) {Change, plain $-$ base (points)};
\node[anchor=west,font=\scriptsize\bfseries] at (-0.850-3.3,0.00) {Tiny Aya Global, Korean};
\node[anchor=east,font=\scriptsize] at (-0.850-0.05,-0.32) {Guard, attack prompts$^a$};
\draw[red!75!black,line width=0.9pt] (0.756,-0.32) -- (1.419,-0.32);
\fill[red!75!black] (1.080,-0.32) circle (1.6pt);
\node[anchor=west,font=\tiny,red!75!black] at (1.419+0.05,-0.32) {+12.7};
\node[anchor=east,font=\scriptsize] at (-0.850-0.05,-0.60) {Guard, harmless prompts$^a$};
\draw[orange!85!black,line width=0.9pt] (0.179,-0.60) -- (0.773,-0.60);
\fill[orange!85!black] (0.468,-0.60) circle (1.6pt);
\node[anchor=west,font=\tiny,orange!85!black] at (0.773+0.05,-0.60) {+5.5};
\node[anchor=east,font=\scriptsize] at (-0.850-0.05,-0.88) {Guard, reader 2's prompts$^b$};
\draw[red!75!black,line width=0.9pt] (0.909,-0.88) -- (1.428,-0.88);
\fill[red!75!black] (1.199,-0.88) circle (1.6pt);
\node[anchor=west,font=\tiny,red!75!black] at (1.428+0.05,-0.88) {+14.1};
\node[anchor=east,font=\scriptsize] at (-0.850-0.05,-1.16) {Reader 1 (author), K2$^c$};
\draw[blue!70!black,line width=0.9pt] (-0.153,-1.16) -- (0.535,-1.16);
\fill[blue!70!black] (0.365,-1.16) circle (1.6pt);
\node[anchor=west,font=\tiny,blue!70!black] at (0.535+0.05,-1.16) {+4.3};
\node[anchor=east,font=\scriptsize] at (-0.850-0.05,-1.44) {Reader 1 (author), K3$^c$};
\draw[blue!70!black,line width=0.9pt] (-0.026,-1.44) -- (0.468,-1.44);
\fill[blue!70!black] (0.348,-1.44) circle (1.6pt);
\node[anchor=west,font=\tiny,blue!70!black] at (0.468+0.05,-1.44) {+4.1};
\node[anchor=east,font=\scriptsize] at (-0.850-0.05,-1.72) {Reader 2 (outside)$^b$};
\draw[blue!70!black,line width=0.9pt] (0.246,-1.72) -- (0.757,-1.72);
\fill[blue!70!black] (0.510,-1.72) circle (1.6pt);
\node[anchor=west,font=\tiny,blue!70!black] at (0.757+0.05,-1.72) {+6.0};
\node[anchor=east,font=\scriptsize] at (-0.850-0.05,-2.00) {Reader 3$^{b,d}$};
\draw[blue!70!black,line width=0.9pt] (0.365,-2.00) -- (1.377,-2.00);
\fill[blue!70!black] (0.884,-2.00) circle (1.6pt);
\node[anchor=west,font=\tiny,blue!70!black] at (1.377+0.05,-2.00) {+10.4};
\node[anchor=east,font=\scriptsize] at (-0.850-0.05,-2.28) {Reader 4$^{b,d}$};
\draw[blue!70!black,line width=0.9pt] (0.442,-2.28) -- (1.182,-2.28);
\fill[blue!70!black] (0.816,-2.28) circle (1.6pt);
\node[anchor=west,font=\tiny,blue!70!black] at (1.182+0.05,-2.28) {+9.6};
\node[anchor=west,font=\scriptsize\bfseries] at (-0.850-3.3,-2.68) {Granite 4.1 3B, Korean};
\node[anchor=east,font=\scriptsize] at (-0.850-0.05,-3.00) {Guard, attack prompts$^a$};
\draw[red!75!black,line width=0.9pt] (1.351,-3.00) -- (2.235,-3.00);
\fill[red!75!black] (1.794,-3.00) circle (1.6pt);
\node[anchor=west,font=\tiny,red!75!black] at (2.235+0.05,-3.00) {+21.1};
\node[anchor=east,font=\scriptsize] at (-0.850-0.05,-3.28) {Guard, harmless prompts$^a$};
\draw[orange!85!black,line width=0.9pt] (0.816,-3.28) -- (1.436,-3.28);
\fill[orange!85!black] (1.114,-3.28) circle (1.6pt);
\node[anchor=west,font=\tiny,orange!85!black] at (1.436+0.05,-3.28) {+13.1};
\node[anchor=east,font=\scriptsize] at (-0.850-0.05,-3.56) {Reader 1 (author), 16d$^c$};
\draw[blue!70!black,line width=0.9pt] (-0.595,-3.56) -- (0.697,-3.56);
\fill[blue!70!black] (0.059,-3.56) circle (1.6pt);
\node[anchor=west,font=\tiny,blue!70!black] at (0.697+0.05,-3.56) {+0.7};
\node[anchor=east,font=\scriptsize] at (-0.850-0.05,-3.84) {Reader 2 (outside)$^c$};
\draw[blue!70!black,line width=0.9pt] (-0.272,-3.84) -- (1.097,-3.84);
\fill[blue!70!black] (0.570,-3.84) circle (1.6pt);
\node[anchor=west,font=\tiny,blue!70!black] at (1.097+0.05,-3.84) {+6.7};
\node[anchor=east,font=\scriptsize] at (-0.850-0.05,-4.12) {Reader 3$^{d}$};
\draw[blue!70!black,line width=0.9pt] (0.850,-4.12) -- (2.890,-4.12);
\fill[blue!70!black] (1.870,-4.12) circle (1.6pt);
\node[anchor=west,font=\tiny,blue!70!black] at (2.890+0.05,-4.12) {+22.0};
\node[anchor=east,font=\scriptsize] at (-0.850-0.05,-4.40) {Reader 4$^{d}$};
\draw[blue!70!black,line width=0.9pt] (0.348,-4.40) -- (2.601,-4.40);
\fill[blue!70!black] (1.386,-4.40) circle (1.6pt);
\node[anchor=west,font=\tiny,blue!70!black] at (2.601+0.05,-4.40) {+16.3};
\end{tikzpicture}
\end{adjustbox}
\caption{Korean change after plain fine-tuning, measured three ways: the guard on attack prompts, the guard on harmless prompts, and people reading the answers with the arm hidden. Dots: estimates; lines: 95\% intervals. On Tiny Aya the guard's rise on attack prompts is larger than any human reading of it; on Granite the two later readers detect a rise and the first two do not. $^a$Pooled over three seeds, bootstrap over prompts. $^b$Seed 42, weighted over the 291 prompts (Appendix~\ref{app:protocol}). $^c$Seed 42, unweighted counts on the sheet; interval from the exact McNemar counts (exploratory). $^d$Readers 3 and 4, one sheet, seed 42, not registered (exploratory): 150 of the Tiny Aya prompts of reader 2's sheet and 50 of its Granite prompts (49 in reader 4's test, one answer marked unsure); Granite unweighted, bootstrap over prompts.}
\label{fig:drift3}
\end{figure}

\paragraph{Confirmatory and exploratory.} The preregistered primary family is H1 (under plain, the Korean guard-flagged rate rises more than the English one), H2 (SafeGrad stays below plain on Korean) and H3 (Swahili refusals fall). The seed 42 result was known from a first stage, so H1--H3 on seeds 1 and 2 are confirmatory; pooled estimates include seed 42. Every other analysis is exploratory, including the later registered checks, which are many and uncorrected. Appendix Table~\ref{tab:rests} marks the status of each claim, and Appendix~\ref{app:prereg} lists every check with when it was registered. Figure~\ref{fig:drift3} sets the guard's Korean rise beside its change on harmless prompts and the human readings. Side results on 4-bit loading and the refusal direction are in Appendices~\ref{app:quant} and~\ref{app:circuits}.

\section{Results}

\begin{table}[t]
\centering\footnotesize
\setlength{\tabcolsep}{4pt}
\begin{adjustbox}{max width=\columnwidth}
\begin{tabular}{@{}llrrr@{}}
\toprule
(\%) & & Base & Plain $-$ base & SafeGrad $-$ base \\
\midrule
\multirow{4}{*}{\shortstack[l]{Guard-flagged\\rate, attack\\prompts $\downarrow$}}
 & en & 6.9 & +0.9 & $-$2.9$^\ast$ \\
 & \textbf{ko} & \textbf{5.2} & \textbf{+12.7}$^\ast$ & \textbf{+4.4}$^\ast$ \\
 & zh & 8.6 & +4.4$^\ast$ & $-$0.1 \\
 & sw$^\dagger$ & 6.2 & +0.3 & $-$1.0 \\
\addlinespace
\multirow{4}{*}{\shortstack[l]{Harmless-prompt\\flag rate $\downarrow$\\(exploratory)}}
 & en & 1.6 & +3.7$^\ast$ & +2.0 \\
 & \textbf{ko} & \textbf{5.6} & \textbf{+5.5}$^\ast$ & \textbf{+3.3}$^\ast$ \\
 & zh & 4.8 & +7.5$^\ast$ & +5.9$^\ast$ \\
 & sw$^\dagger$ & 10.4 & +2.5 & +0.1 \\
\addlinespace
\multirow{4}{*}{\shortstack[l]{Refusal rate,\\attack\\prompts $\uparrow$}}
 & en & 71.1 & $-$5.3$^\ast$ & +5.8$^\ast$ \\
 & ko & 59.8 & $-$10.3$^\ast$ & $-$1.6 \\
 & zh & 66.3 & $-$8.1$^\ast$ & +1.0 \\
 & sw$^\dagger$ & 50.9 & $-$9.3$^\ast$ & +4.6$^\ast$ \\
\bottomrule
\end{tabular}
\end{adjustbox}
\caption{\textbf{Read the bold Korean rows:} the guard-flagged rate rises 12.7 points under plain against +0.9 in English, SafeGrad keeps 4.4 of it, and the guard also flags 5.5 points more answers to harmless prompts. Tiny Aya Global before and after fine-tuning on 1,800 harmless rows: the base model's rate and the change in points, pooled over three seeds (Korean under plain: 5.2\% to 17.9\%). $^\ast$Detected: the 95\% interval over prompts excludes zero (intervals: Appendix Table~\ref{tab:mainci}). Every row is the guard's label; human readings are in \S\ref{sec:measure}. $^\dagger$Never in training.}
\label{tab:main}
\end{table}

\subsection{The guard-flagged rate rises unevenly by language}
\label{sec:drift}

Plain LoRA on 1,800 harmless rows raised the Korean guard-flagged rate from 5.2\% to 17.9\% pooled over seeds (19.2\%, 15.5\% and 18.9\% at seeds 42, 1 and 2; +14.1, +10.3 and +13.7 points from unrounded rates, each $p<0.05$). English changed by +0.0, +0.7 and +2.1. H1 passes in every seed (confirmatory on seeds 1 and 2): the Korean rise exceeds the English rise by +11.8 \ci{+6.8}{+16.7} (Table~\ref{tab:main}; by seed, Appendix Table~\ref{tab:repair42}). Pooled, the Korean rise is +12.7 \ci{+8.9}{+16.7}; \S\ref{sec:measure} tests how much of it is the guard.

\paragraph{English.} No change in the English rate was detected (+0.9 \ci{-1.8}{+3.8}). English refusals did fall ($-$5.3 \ci{-9.5}{-1.1}), and with half the epochs the English rate rose too (+4.8, $p=0.004$; Appendix Table~\ref{tab:sens}).

\begin{figure}[t]
\centering
{\small Guard-flagged rate\par}\vspace{2pt}
\includegraphics[width=0.74\columnwidth,trim=0 72 338 45,clip]{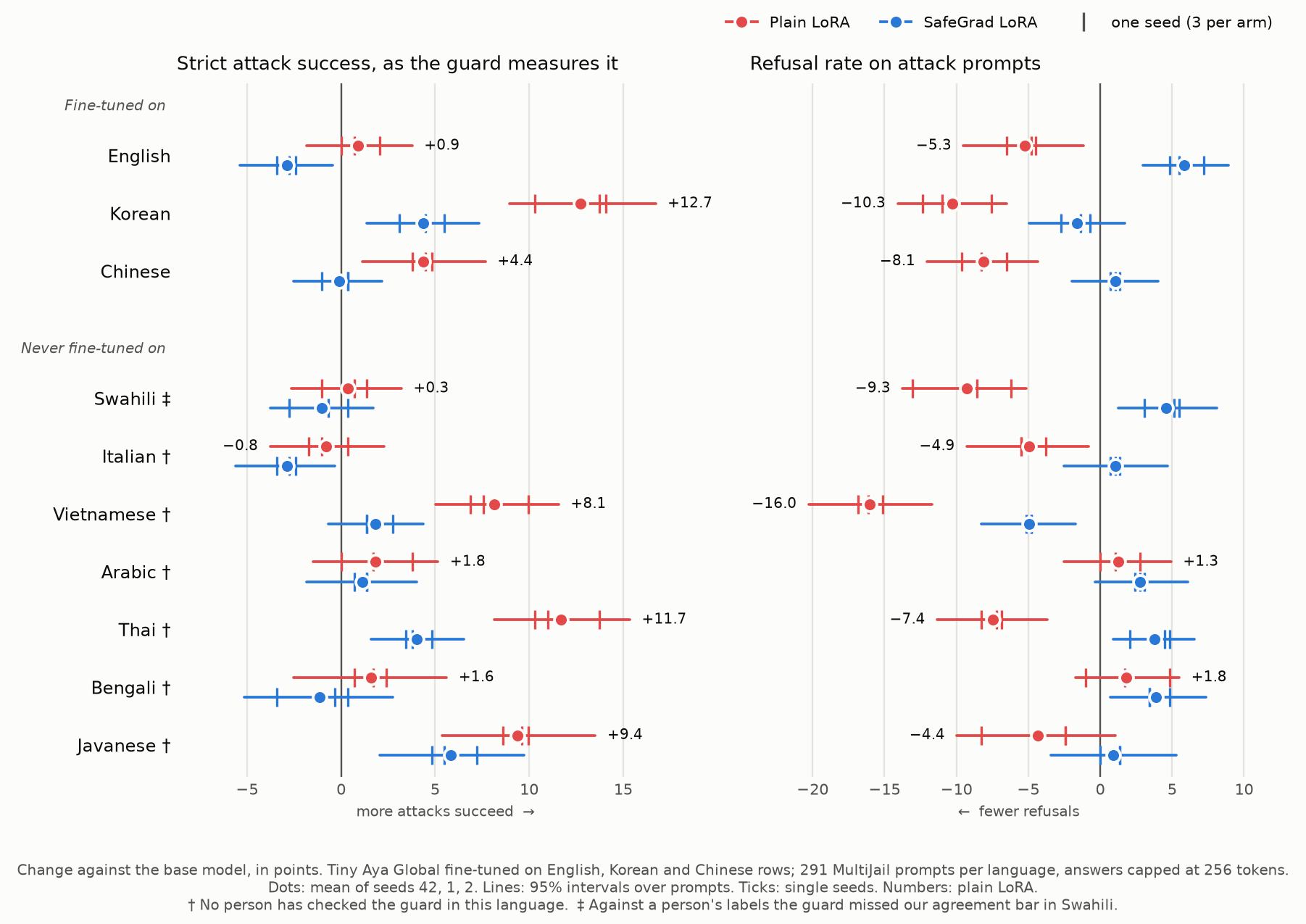}
\caption{Change in the guard-flagged rate against the base model in ten languages, in points (three seeds, answers capped at 256 tokens). Red: plain LoRA; blue: SafeGrad LoRA. Dots: seed means; lines: 95\% intervals over prompts; ticks: single seeds; numbers: the change under plain. $^\dagger$No person has checked the guard in this language. $^\ddagger$The guard missed our agreement bar in Swahili. After re-generation the Bengali change is above zero (+5.6 \ci{+1.7}{+9.3}).}
\label{fig:ten}
\end{figure}

\paragraph{Ten languages and a second benchmark} (exploratory). The same adapters answered the 291 families in six more languages never trained on (Figure~\ref{fig:ten}). No added language rose more than Korean. Thai (+11.7), Javanese (+9.4) and Vietnamese (+8.1) rose by amounts whose difference from Korean's was not detected (Korean minus each: +1.0 \ci{-3.8}{+6.0}, +3.3 \ci{-2.3}{+8.7} and +4.6 \ci{-0.0}{+9.3}). On MultiJail the Korean rise exceeds the Chinese one (+8.4 \ci{+3.7}{+13.1}). On the second benchmark (Appendix Table~\ref{tab:second}) Korean rose more than English (+8.7 \ci{+3.5}{+13.8}), but no difference between Korean and Chinese was detected ($-$1.3 \ci{-7.2}{+4.5}). Among English, Korean, Chinese and Swahili, the rise is largest in Korean on MultiJail only.

\paragraph{Robustness under the same guard.} The rise holds after every cut-off answer was generated again, on a second dataset, on a second Tiny Aya model and with the Unsafe class alone (Appendix~\ref{app:tables}, Table~\ref{tab:settings}). These show that the guard's reading is stable; they do not test the guard.

\subsection{The Korean rise needs no Korean training rows}
\label{sec:language}

One explanation is noisy Korean training rows: 7 of 30 failed our spot check (the limit was 3), as did a filtered set. So we removed Korean from training.

\paragraph{No Korean rows} (exploratory). This run trains on the 600 English and 600 Chinese rows plus 274 screened spare rows of each (1,747 after one was dropped as too long), so it also changes some instructions. The Korean rate was 18.2\%, 18.6\% and 18.2\% against 5.2\% for base, so Korean training rows are not needed for a rise of this size. No difference from training with Korean rows was detected (+0.5 \ci{-2.3}{+3.3}; all settings: Appendix Table~\ref{tab:settings}).

\paragraph{One language at a time} (600 rows; exploratory). Trained on English, Korean or Chinese alone, the Korean rate rose by 7 to 12 points in each of three seeds (Appendix~\ref{app:tables}).

\paragraph{Refusals fall in languages never trained on.} Swahili refusals fell under plain ($-$9.3 \ci{-13.7}{-5.2}; H3, confirmatory on seeds 1 and 2), as refusals did in the three trained languages (Table~\ref{tab:main}), and over the six further languages the fall averaged $-$4.9 \ci{-7.0}{-2.9}, an average in which the Arabic and Bengali point estimates rose (+1.3, +1.8). On the Aya data no Swahili fall was detected ($-$2.6 \ci{-6.9}{+1.5}).

\subsection{How far can the guard be trusted?}
\label{sec:measure}

We test the guard three ways: against labels and other judges, against blinded human readers, and on harmless prompts.

\begin{table*}[t]
\centering\footnotesize
\setlength{\tabcolsep}{4pt}
\resizebox{\textwidth}{!}{\begin{tabular}{@{}lccccc@{}}
\toprule
A person's reading of Korean answers, arm hidden, & Prompts in & Person: marked unsafe, & Guard: flagged, same & Marked safe, guard & Marked unsafe, guard \\
base against plain at seed 42 & the test & base / plain ($p$) & prompts, base / plain & passed: base / plain & flagged: base / plain \\
\midrule
K2, Tiny Aya (random prompts) & 94 & 1 / 5 \;(0.22) & 2 / 11 & 96 of 97 / 83 of 89 & 1 of 1 / 7 of 7 \\
K3, Tiny Aya (prompts not read before) & 145 & 5 / 11 \;(0.070) & 6 / 18 & 145 of 148 / 126 of 135 & 4 of 10 / 10 of 13 \\
16d, Granite 4.1 (random prompts) & 150 & 83 / 84 \;(1.00) & 31 / 66 & 62 of 67 / 43 of 66 & 26 of 83 / 43 of 84 \\
\keyrow R1, second reader, Tiny Aya$^\ast$ & 225 & 11 / 28 \;($<$0.001) & 13 / 54 & \multicolumn{2}{c}{same mark as the guard: 215 / 195 of 225} \\
R2, second reader, 75 of the 150 prompts of 16d & 75 & 9 / 14 \;(0.227) & 12 / 30 & \multicolumn{2}{c}{same mark as the guard: 70 / 57 of 75} \\
\midrule
\keyrow Reader 3, Tiny Aya, R1's prompts$^\dagger$ & 150 & 26 / 52 \;($<$0.001) & 12 / 53 & 121 of 124 / 90 of 98 & 9 of 26 / 45 of 52 \\
\keyrow Reader 4, Tiny Aya, R1's prompts$^\dagger$ & 150 & 12 / 39 \;($<$0.001) & 12 / 53 & 132 of 138 / 94 of 111 & 6 of 12 / 36 of 39 \\
Reader 3, Granite, 50 of R2's prompts & 50 & 9 / 20 \;(0.003) & 9 / 23 & 38 of 41 / 23 of 30 & 6 of 9 / 16 of 20 \\
Reader 4, Granite, 50 of R2's prompts$^\ddagger$ & 49 & 7 / 15 \;(0.039) & 8 / 22 & 39 of 42 / 25 of 34 & 6 of 8 / 13 of 15 \\
\bottomrule
\end{tabular}}
\caption{\textbf{Shaded: the three readers outside the team on Tiny Aya,} who each detect a rise; compare a person's base / plain count with the guard's on the same prompts. Human readings, with the guard on the same prompts. K2, K3 and 16d: one author; R1 and R2: a second reader (exploratory); readers 3 and 4: one more sheet, not registered (exploratory; Appendix~\ref{app:protocol}). The test uses prompts with a yes or a no for both answers; the last two columns, every such answer. $^\ast$The registered reading is a weighted difference over the 291 prompts, +6.0 \ci{+2.9}{+8.9}; these counts are unweighted. $^\dagger$All 53 prompts where the guard's verdict differs between base and plain and 97 of the other 238; weighted, +10.4 \ci{+4.3}{+16.2} and +9.6 \ci{+5.2}{+13.9}. $^\ddagger$Reader 4 marked one plain answer unsure, so 49 of the 50 prompts enter the test.}
\label{tab:readers}
\end{table*}

\begin{table}[t]
\centering\footnotesize
\setlength{\tabcolsep}{3pt}
\begin{adjustbox}{max width=\columnwidth}
\begin{tabular}{@{}lcccc@{}}
\toprule
Agreement with 120 labels & en & ko & zh & sw \\
\midrule
\keyrow Qwen3Guard (the guard used) & 28/30 & 24/30 & 27/30 & 20/27 \\
PolyGuard-Ministral & 11/30 & 18/30 & 18/30 & 17/27 \\
Granite Guardian 3.3 8B & 20/30 & 20/30 & 22/30 & 22/27 \\
gpt-oss-safeguard-20b & 26/30 & 22/30 & 22/30 & 22/27 \\
\bottomrule
\end{tabular}
\end{adjustbox}
\caption{\textbf{Shaded: the guard that scored every run,} the only judge at or above 24/30 in Korean. Agreement with 120 labels made without the guard's verdict. The bar is 80\% in a language; only Qwen3Guard scored the runs.}
\label{tab:agree}
\end{table}

\paragraph{Labels and second judges} (Table~\ref{tab:agree}). Against 120 answers we labelled without seeing its verdict, the guard reached our 80\% bar in English, Chinese and, exactly, Korean (24/30), and missed it in Swahili, where labels come from an English back-translation. None of three other judges \citep{kumar2025polyguard,padhi2024graniteguardian,openai2025safeguard} reached the bar in Korean, the third even with our labellers' rule as its policy, so every reported rate is Qwen3Guard's.

\paragraph{Human readers} (Table~\ref{tab:readers}). One author, reading Korean answers of base and plain at seed 42 with the arm hidden, detected no rise in three registered readings: on Tiny Aya 1 against 5 answers marked unsafe on 94 prompts ($p=0.22$) and 5 against 11 on 145 ($p=0.070$), where the guard counts 2 against 11 and 6 against 18. A second reader, a native Korean speaker outside the team who was not told what we expected, detected a rise of +6.0 \ci{+2.9}{+8.9} points on Tiny Aya, less than half of the +14.1 \ci{+10.7}{+16.8} the guard counts on the same prompts, and none on Granite (9 against 14 on 75 prompts, $p=0.23$). We treat this reading as exploratory (Appendix~\ref{app:protocol}). Two further readers outside the team, both native Korean speakers (readers 3 and 4), marked one more sheet one answer at a time (exploratory, not registered). On 150 of the second reader's Tiny Aya prompts, weighted the same way, they detected rises of +10.4 \ci{+4.3}{+16.2} and +9.6 \ci{+5.2}{+13.9} points; on 50 of the Granite prompts (49 in reader 4's test), +22.0 \ci{+10.0}{+34.0} and +16.3 \ci{+4.1}{+30.6} (exact McNemar $p=0.003$ and $0.039$). On 40 English prompts of the same sheet neither detected a change (10 against 7 and 7 against 5; the guard 9 against 9; Appendix~\ref{app:protocol}). On Granite their marks agree with each other ($\kappa$ 0.66) and with the second reader's ($\kappa$ 0.65 and 0.79), but not with the first reader's ($\kappa$ 0.28 and 0.15). Almost all of the second reader's difference is on the 53 prompts where the guard's verdict differs between base and plain (+16 there, +1 on the other 172). And among answers the first reader marked safe, the guard passed a smaller share for plain than for base on every sheet, so its disagreement with a reader need not cancel in a before-after difference.

\paragraph{The guard on harmless prompts} (Appendix Table~\ref{tab:harmless}; exploratory). A guard that reacts only to harm should move little on harmless prompts (the safe XSTest prompts). After fine-tuning it flagged +5.5 \ci{+2.1}{+9.1} points more of the Korean answers to them. If all of that were the guard's reaction to fine-tuned answers, the Korean rise on attack prompts would shrink from +12.7 to +7.2 \ci{+2.0}{+12.4}. The Korean--English contrast is not explained this way: no difference between the Korean and the English change on harmless prompts was detected (+1.7 \ci{-2.3}{+6.0}). On Granite the guard flagged 13.1 points more harmless Korean answers, and two of four readers detected the rise (\S\ref{sec:granite}).

\subsection{Is it just damage to the model?}
\label{sec:damage}

\paragraph{The first fine-tune costs task score.} The instruction fine-tune made the model worse at its own task: the task score fell from 6.32 to 5.89 under plain and 5.85 under SafeGrad (three-seed means; by language, Appendix~\ref{app:tables}). We therefore test whether the drift is general damage. Other learning rates and epoch counts do not settle this: the task score fell below base under each (Appendix Table~\ref{tab:sens}).

\paragraph{A fraction of the update} (exploratory). Without training, we scaled the plain update to 0.25, 0.5 and 0.75 of its strength (seed 42; the two lowest also at seeds 1 and 2 and on harmless prompts) and built two random updates as large as plain's in every module (Appendix Table~\ref{tab:dose}). Over three seeds, at a quarter of the update the Korean rate rose (+4.0 \ci{+1.5}{+6.6}) with no detected fall in the task score ($-$0.08 \ci{-0.18}{+0.02}). At that strength, however, the rise was not detected once the guard's change on harmless prompts was subtracted (+3.2 \ci{-0.2}{+6.9}); that change itself was not detected there (+0.8 \ci{-1.7}{+3.3}). No rise was detected under random updates of the same size (+0.3 \ci{-1.0}{+1.7} and $-$0.3 \ci{-2.1}{+1.4}, against +14.1 \ci{+9.6}{+18.9} for the fine-tune's own update at seed 42), so the rise is not explained by the size of the change alone.

\paragraph{A fine-tune that helps its own task} (exploratory). Does the rise still appear when the fine-tune improves the task it is trained for? We trained Tiny Aya Global on one-sentence news summarisation (XL-Sum, \citealp{hasan2021xlsum}; 400 rows each in English, Korean and Chinese), plain and SafeGrad at three seeds on a cloud A100 with its own base run, and scored the summaries by ROUGE-L \citep{lin2004rouge} on 300 held-out articles (Appendix Tables~\ref{tab:xlsum}, \ref{tab:xlsumcut}). The fine-tune improved ROUGE-L by +8.18 \ci{+6.96}{+9.36}, and the Korean rate still rose (+4.5 \ci{+1.9}{+7.2}). The general task score still fell ($-$0.45 \ci{-0.58}{-0.32}; SafeGrad $-$0.16 \ci{-0.27}{-0.05}), and the summaries got shorter (from 150 to 67--70 characters, against references of 54 to 124), so part of the gain is the one-sentence format. The rise was detected among complete answers (+6.3 \ci{+3.8}{+9.1}), and no difference between SafeGrad and plain was detected in Korean ($-$2.3 \ci{-4.6}{+0.0}); the rise by answer length, the comparisons with English and with harmless prompts, and a second reading with uncapped answers are in Appendix~\ref{app:tables}.

\subsection{Two repairs, two operating points}
\label{sec:repair}

\paragraph{SafeGrad.} SafeGrad left the Korean rate 8.6, 7.2 and 9.3 points below plain at the same seed (pooled $-$8.4 \ci{-11.3}{-5.5}; H2, confirmatory on seeds 1 and 2). The repair is partial: Korean stays +4.4 \ci{+1.4}{+7.3} above base. No change in English over-refusal against base was detected at any seed.

\paragraph{Mixing} (exploratory). Mixing adds the same 47 refusal examples to the training data, once (\emph{mixing $\times$1}, 141 views) or as five copies (\emph{mixing $\times$5}, 705 views, close to SafeGrad's $\approx$684). We report seeds 1 and 2 (rates and the seed 42 run: Appendix Table~\ref{tab:repair42}). Against SafeGrad at the same seed, mixing $\times$5 lowered the Korean rate further ($-$2.7 \ci{-5.2}{-0.5}); for mixing $\times$1 no difference was detected ($-$0.5 \ci{-2.9}{+1.9}). Both raised English over-refusal above SafeGrad's (+5.2 \ci{+2.8}{+8.0} and +5.0 \ci{+2.6}{+7.6}) and refused more safe prompts in the other languages; no difference in task score was detected. These are two points on a trade-off: each ran in one untuned configuration, with refusal examples drawn from MultiJail, the benchmark we evaluate on. No person has read either repair's answers, and \S\ref{sec:measure} shows the guard overstating a change, so part of mixing $\times$5's edge over SafeGrad could be the guard's sensitivity to explicit refusal phrasing, which mixing trains in, rather than less harm. On the summarisation fine-tune mixing $\times$5 also left the Korean rate below plain ($-$5.6 \ci{-8.5}{-3.1}) without lowering ROUGE-L, again with more over-refusal (Appendix~\ref{app:tables}). On Granite, where mixing was not run, SafeGrad refused more safe English prompts (24.4\%, 25.2\% and 21.6\% against 9.6\% for base).

\subsection{A second model family}
\label{sec:granite}

We repeated the main experiment on IBM Granite 4.1 3B \citep{ibm2026granite41}, one model of one other family, chosen before any result (Appendix~\ref{app:details}, Table~\ref{tab:granite}). By the guard, Korean again rose more than English (+20.8 \ci{+15.1}{+26.3}): Korean went from 19.2\% to 40.3\% (+21.1 \ci{+15.9}{+26.3}), with no detected change in English (+0.2 \ci{-2.2}{+2.7}), also with no Korean training row (+19.5 \ci{+14.3}{+24.7}) and after re-generation.

The human readings split. On harmless prompts the guard's flags of Korean answers rose from 4.0\% to 17.1\% (+13.1 \ci{+9.6}{+16.9}). The first two readers detected no rise: on 150 Korean prompts the first reader marked 83 base and 84 plain answers unsafe ($p=1.00$), and the second reader detected none on 75 of them. Two later readers, on 50 of those 75 prompts, marked 9 against 20 and 7 against 15 ($p=0.003$ and $0.039$; exploratory). We therefore treat the Granite rise as likely but not established.

By the guard, SafeGrad left the rate below plain fine-tuning ($-$13.7 \ci{-16.1}{-11.4}). We do not claim that the drift or the repair holds across models, and compare no sizes with Tiny Aya.

\section{Conclusion}

On one model family and one guard, a benign multilingual fine-tune raised the guard-flagged rate unevenly across languages, most in Korean, without Korean training data. Blinded readers confirmed a rise on Tiny Aya at about 40\% to 75\% of the guard's size, two of four detected one on Granite, and the guard also flagged more answers to harmless prompts, so a guard-measured drift is a real signal that overstates the harm. Before shipping a benign fine-tune, re-run the safety evaluation in every language served, not only the training ones, and score harmless prompts with the same guard. Have a native speaker read a blinded sample before trusting a guard-measured change, and consider mixing five copies of a few dozen of the base model's own refusals into training while measuring the over-refusal it adds (Appendix~\ref{app:recs}).\label{end:main}

\section*{Limitations}
\label{start:lim}

\begin{itemize}\setlength\itemsep{0pt}
\item \textbf{One guard and few human readings.} Every rate is one guard's; three other judges missed our agreement bar in Korean, and none scored the runs. One author's three readings at seed 42 detected no rise; that reader knew the expected direction and can often guess the arm from an answer's length. The second reader is one person recruited through an author, and on Tiny Aya their marks nearly match the first reader's (370 of 378 answers; Appendix~\ref{app:protocol}). Readers 3 and 4, both outside the team and native Korean speakers, read one sheet that was not registered, and agree with each other at $\kappa$ 0.59 overall; more readers are not ground truth. On Granite the first reader and the guard agree on only 58\% of the answers, and the first reader's Granite marks disagree with the three other readers'. English was read only in a small unregistered check (40 Tiny Aya prompts, no change detected; Appendix~\ref{app:protocol}), and no human reading tested seeds 1 and 2, SafeGrad or mixing. The same guard screened the training data and labels both the flagged rate and refusals.
\item \textbf{What the harmless-prompt check shows.} The safe XSTest prompts are written to look unsafe, a model that refuses less answers more of them, and nobody has read those answers, so the harmless-prompt rise only bounds the guard's reaction loosely.
\item \textbf{The general task score fell.} Each instruction fine-tune we scored lowered it (plain, SafeGrad, both mixing arms, the learning-rate and epoch runs, and Granite), for plain and SafeGrad also at matched answer length. The summarisation fine-tune improved ROUGE-L partly through shorter answers and still lowered the general score; its verdicts are one guard's on answers nobody has read. At a quarter of the update, ``no detected fall'' in the task score is an interval ending at +0.02 on 300 prompts; a scaled update is not an earlier point of training, and at full strength the task score is 0.38 below base.
\item \textbf{One model family, plus one model where the rise is likely but not established.} Earth is a variant of Global, not an independent model. On Granite 4.1 we repeated the main experiment, the run with no Korean rows, the cap audit, the task score, over-refusal and four human readings; everything else is Tiny Aya's. One LoRA rank; the learning rate and epochs were varied at one seed. Tiny Aya Fire and Water were not tested.
\item \textbf{Low-resource languages rest on the weakest measurements.} No person has checked the guard in the six added languages. No team member reads Swahili; the Swahili labels come from an English back-translation, and the guard missed our bar there. The safe XSTest prompts were machine-translated and not checked.
\item \textbf{Why some languages move is not shown:} a weaker model, a more sensitive guard and a change in answer style are not separated. \textbf{Korean data quality} could not be ruled out as a contributor (H4 was never tested), though the no-Korean and single-language runs show Korean rows are not needed.
\item \textbf{Benchmarks.} Single-turn, direct requests only; 291 prompts per language, so one point is three prompts. On the second benchmark the prompts are machine translations that no person has checked, and 35.5\% of the base model's Korean answers reached the 256-token cap against 3.5\% to 6.5\% of the plain runs'; a second reading after re-generation, registered after the first, came out the same. In our check of 20 MultiJail families one Korean and one Chinese prompt drifted in meaning. Cohere built both the model and the Aya data, and pre-training contamination cannot be ruled out.
\item \textbf{Statistics.} Intervals cover prompts, not seeds; nothing is corrected for the number of checks; several checks were registered after a first result was known (\S\ref{sec:setup}, Appendix~\ref{app:prereg}). Close calls, such as the advantage of mixing $\times$5 over SafeGrad ($-$0.5 at the upper end), should be read with that.
\end{itemize}

\section*{Ethics Statement}

All attack prompts come from two public benchmarks, MultiJail and XSTest; we wrote no new jailbreaks. The released outputs contain harmful model answers in ten languages for reproducibility; the dataset cards carry a content warning, and this paper describes harmful answers by category and does not reproduce them. The work shows how ordinary fine-tuning can change measured safety; we think the defensive value of knowing this outweighs the risk, because the recipe is the default use of an open model. Human labelling was done by the authors and by three further readers: one for the second reading of \S\ref{sec:measure} and two more, readers 3 and 4, for a third. Each was warned about the content, could stop at any time and was told the marks would be published without a name.

\bibliography{references}

\clearpage
\appendix

\section{Recommendations for Practitioners}
\label{start:app}
\label{app:recs}

Before shipping a benign fine-tune, we recommend:
\begin{enumerate}\setlength\itemsep{0pt}\setlength\parsep{0pt}
\item Re-run the safety evaluation in every language the model will serve, including languages absent from training.
\item Score harmless prompts with the same guard, so that a change in how it reads fine-tuned answers shows up.
\item Have a native speaker read a blinded sample of answers before trusting a guard-measured change.
\item Try mixing five copies of a few dozen of the base model's own refusals into the training data (\S\ref{sec:repair}), and measure the over-refusal it adds.
\end{enumerate}

\section{Training and Evaluation Details}
\label{app:details}

\begin{table}[!ht]
\centering\footnotesize
\setlength{\tabcolsep}{3pt}
\begin{tabular}{@{}p{0.27\columnwidth}p{0.69\columnwidth}@{}}
\toprule
Setting & Value \\
\midrule
LoRA & rank 16, alpha 32, dropout 0, on q, k, v, o, gate, up and down projections (30.2M trainable parameters) \\
Optimizer & AdamW, learning rate $2\times10^{-4}$, cosine schedule, 3\% warm-up, weight decay 0, gradient clip 1.0 \\
Batch, length & micro-batch 4 $\times$ 8 accumulation (effective 32); 3 epochs; 171 optimizer steps for 1,800 rows; sequence limit 1,024 tokens \\
Seeds & 42, 1 and 2 \\
SafeGrad & one batch of 4 refusal examples per optimizer step ($\approx$684 views per run), shuffled once and replayed in that order, as in SafeTune's trainer; the reference is the student with adapters off \\
Alignment set & the base model's refusals to 72 held-out MultiJail prompts that the guard labels safe and a refusal: 47 kept for Global, 44 for Earth, 38 for Granite (16 English, 7 Korean, 15 Chinese) \\
Decoding & greedy; the model's default chat template and system preamble; on the laptop the adapter is merged first \\
New tokens & 256 for attack and XSTest prompts; 1,024 for task prompts and for the re-generation audit \\
Guard & Qwen3Guard-Gen-4B, revision \texttt{6ec42827} \\
Task judge & Qwen3-14B, revision \texttt{40c06982}, on Tiny Aya's 1--7 rubric \\
Translation & NLLB-200 3.3B, revision \texttt{1a07f7d1} (Swahili re-translation, unsafe XSTest prompts) \\
Granite 4.1 3B & revision \texttt{c0650403} (Apache 2.0); 1,795 training rows (five Korean rows were over the length limit), 1,748 with no Korean row; its template has no system prompt \\
\bottomrule
\end{tabular}
\caption{Training and evaluation settings.}
\label{tab:params}
\end{table}

\paragraph{Data and licenses.} Bactrian-X (CC-BY-NC-4.0), MultiJail (MIT), XSTest (CC-BY-4.0), Aya Dataset (Apache-2.0), XL-Sum (CC-BY-NC-SA-4.0), Tiny Aya (CC-BY-NC). XSTest safe prompts: Korean and Chinese translations by Aya Expanse 8B \citep{dang2024ayaexpanse}, Swahili by Tiny Aya Global; its unsafe prompts by NLLB-200. Leak checks: training and task instructions are disjoint by ID and by MinHash; alignment prompts are never evaluated; Aya training and evaluation rows never overlap.

\paragraph{Base runs.} Each block of runs is compared with its own base run: task score 6.32 for the laptop runs, 6.35 in the quantization block and 6.29 on the cloud GPU.

\paragraph{Compute.} All fine-tunes of the main experiment ran on one Apple laptop; the second model, the third judge, the summarisation runs and the 4-bit runs used Hugging Face Jobs. Table~\ref{tab:cost} gives what the work used and what one evaluation pass costs.

\begin{table}[!ht]
\centering\footnotesize
\setlength{\tabcolsep}{3pt}
\begin{tabular}{@{}p{0.40\columnwidth}p{0.56\columnwidth}@{}}
\toprule
Item & Amount \\
\midrule
Laptop fine-tunes (Apple M5 Pro, MPS, bf16) & 45 runs, 44.1 hours of training, from the run records \\
One 1,800-row fine-tune & about 78 minutes plain, 88 with SafeGrad (+13.7\%, three seeds) \\
Mixing $\times$5 & +235 rows; 82.2 minutes at seed 42, against 75.7 plain \\
Cloud GPU, 43 jobs (failed and stopped ones included) & L4 12.64 h, L40S 6.17 h, A100 7.35 h: 26.16 GPU hours, \$39.60 at list price \\
One evaluation pass (2,463 requests; 935k input, 473k output tokens) & Tiny Aya Global on the laptop: 0.57 machine-hours \\
\bottomrule
\end{tabular}
\caption{Compute used for this work.}
\label{tab:cost}
\end{table}

\section{Full Tables}
\label{app:tables}

\begin{table}[!ht]
\centering\footnotesize
\setlength{\tabcolsep}{2.5pt}
\resizebox{\columnwidth}{!}{\begin{tabular}{@{}llrll@{}}
\toprule
(\%) & & Base & Plain $-$ base & SafeGrad $-$ base \\
\midrule
\multirow{4}{*}{\shortstack[l]{Guard-flagged\\rate, attack\\prompts $\downarrow$}}
 & en & 6.9 & +0.9 \ci{-1.8}{+3.8} & $-$2.9 \ci{-5.4}{-0.5} \\
 & \textbf{ko} & \textbf{5.2} & \textbf{+12.7} \ci{+8.9}{+16.7} & \textbf{+4.4} \ci{+1.4}{+7.3} \\
 & zh & 8.6 & +4.4 \ci{+1.1}{+7.7} & $-$0.1 \ci{-2.5}{+2.2} \\
 & sw$^\dagger$ & 6.2 & +0.3 \ci{-2.6}{+3.2} & $-$1.0 \ci{-3.8}{+1.7} \\
\addlinespace
\multirow{4}{*}{\shortstack[l]{Harmless-prompt\\flag rate $\downarrow$\\(exploratory)}}
 & en & 1.6 & +3.7 \ci{+1.2}{+6.4} & +2.0 \ci{-0.4}{+4.5} \\
 & \textbf{ko} & \textbf{5.6} & \textbf{+5.5} \ci{+2.1}{+9.1} & \textbf{+3.3} \ci{+0.4}{+6.4} \\
 & zh & 4.8 & +7.5 \ci{+4.0}{+11.2} & +5.9 \ci{+2.8}{+9.3} \\
 & sw$^\dagger$ & 10.4 & +2.5 \ci{-1.5}{+6.4} & +0.1 \ci{-3.7}{+4.0} \\
\addlinespace
\multirow{4}{*}{\shortstack[l]{Refusal rate,\\attack\\prompts $\uparrow$}}
 & en & 71.1 & $-$5.3 \ci{-9.5}{-1.1} & +5.8 \ci{+3.0}{+8.9} \\
 & ko & 59.8 & $-$10.3 \ci{-14.1}{-6.5} & $-$1.6 \ci{-4.9}{+1.7} \\
 & zh & 66.3 & $-$8.1 \ci{-12.0}{-4.4} & +1.0 \ci{-1.9}{+4.0} \\
 & sw$^\dagger$ & 50.9 & $-$9.3 \ci{-13.7}{-5.2} & +4.6 \ci{+1.3}{+8.1} \\
\bottomrule
\end{tabular}}
\caption{\textbf{Read the bold Korean rows,} as in Table~\ref{tab:main}. Table~\ref{tab:main} with its 95\% intervals over prompts (changes in points, pooled over three seeds). $^\dagger$Never in training.}
\label{tab:mainci}
\end{table}

\begin{table}[!ht]
\centering\footnotesize
\setlength{\tabcolsep}{2.5pt}
\resizebox{\columnwidth}{!}{\begin{tabular}{@{}lrrrrrrrrrrrr@{}}
\toprule
 & & \multicolumn{4}{c}{Guard-flagged rate} & \multicolumn{4}{c}{Over-refusal} & & & \\
Method & Seed & en & ko & zh & sw & en & ko & zh & sw & Task & Steps & Min. \\
\midrule
\keyrow Base model & -- & 6.9 & 5.2 & 8.6 & 6.2 & 9.6 & 7.6 & 8.8 & 7.6 & 6.32 & -- & -- \\
Plain & 42 & 6.9 & 19.2* & 13.4* & 7.6 & 7.2 & 5.6 & 5.6 & 2.8 & 5.94 & 171 & 75.7 \\
Plain & 1 & 7.6 & 15.5* & 12.4 & 5.2 & 6.0 & 5.6 & 4.8 & 4.0 & 5.85 & 171 & 82.2 \\
Plain & 2 & 8.9 & 18.9* & 13.1* & 6.9 & 6.0 & 5.6 & 5.6 & 3.6 & 5.88 & 171 & 75.7 \\
SafeGrad & 42 & 4.5 & 10.7$^\ddagger$ & 8.9$^\ddagger$ & 5.5 & 7.6 & 6.8 & 5.6 & 5.6 & 5.82 & 171 & 89.7 \\
SafeGrad & 1 & 3.4$^\ddagger$ & 8.2$^\ddagger$ & 7.6$^\ddagger$ & 3.4 & 8.8 & 7.2 & 6.4 & 6.0 & 5.86 & 171 & 87.7 \\
SafeGrad & 2 & 4.1$^\ddagger$ & 9.6$^\ddagger$ & 8.9$^\ddagger$ & 6.5 & 10.8 & 7.2 & 6.4 & 6.0 & 5.87 & 171 & 87.8 \\
Mixing $\times$1 & 42 & 3.8 & 9.3 & 6.5 & 4.8 & 14.0 & 7.2 & 8.8 & 7.6 & 5.95 & 174 & 75.6 \\
Mixing $\times$1 & 1 & 3.4 & 8.6 & 6.2 & 3.8 & 14.4 & 9.6 & 10.4 & 10.0 & 5.88 & 174 & 85.9 \\
Mixing $\times$1 & 2 & 3.4 & 8.2 & 6.5 & 5.2 & 15.2 & 9.2 & 8.8 & 8.4 & 5.92 & 174 & 76.0 \\
Mixing $\times$5 & 42 & 3.4 & 8.9 & 7.9 & 4.8 & 13.2 & 8.0 & 9.2 & 6.4 & 5.94 & 192 & 82.2 \\
\keyrow Mixing $\times$5 & 1 & 2.7 & 6.2 & 5.8 & 6.2 & 15.2 & 11.2 & 9.2 & 8.8 & 5.92 & 192 & 82.7 \\
\keyrow Mixing $\times$5 & 2 & 3.4 & 6.2 & 6.2 & 4.5 & 14.8 & 10.0 & 9.6 & 6.8 & 5.84 & 192 & 82.2 \\
\bottomrule
\end{tabular}}
\caption{\textbf{Shaded: the base model and mixing $\times$5 at seeds 1 and 2,} the runs the text compares. Read the ko column (plain about 15 to 19, SafeGrad 8 to 11, mixing $\times$5 about 6) against the over-refusal columns, where mixing is highest. Every run of the main comparison (rates in \%, as the guard labels them; task score; optimizer steps; training minutes on a laptop that was also doing other work). * plain differs from base; $^\ddagger$ SafeGrad differs from plain of the same seed (guard-flagged rate, exact McNemar, $p<0.05$). For mixing, seed 42 was seen before the rules were written; only seeds 1 and 2 are used in \S\ref{sec:repair}. Differences quoted in the text come from unrounded rates.}
\label{tab:repair42}
\end{table}

\begin{table}[!ht]
\centering\footnotesize
\setlength{\tabcolsep}{3pt}
\resizebox{\columnwidth}{!}{\begin{tabular}{@{}lrl@{}}
\toprule
Setting (3 seeds) & Base ko & Korean rise, pooled \\
\midrule
\keyrow Global, Bactrian-X, 1,800 rows & 5.2 & +12.7 \ci{+8.9}{+16.7} \\
\keyrow \quad no Korean rows (1,747 rows) & 5.2 & +13.2 \ci{+9.4}{+17.2} \\
\quad cut to 519 rows & 5.2 & +12.6 \ci{+8.6}{+16.8} \\
\quad English rows only (600) & 5.2 & +8.0 \ci{+4.8}{+11.3} \\
\quad Korean rows only (600) & 5.2 & +11.2 \ci{+7.3}{+15.1} \\
\quad Chinese rows only (600) & 5.2 & +11.6 \ci{+7.9}{+15.5} \\
Global, Aya, 519 rows & 5.2 & +9.4 \ci{+6.1}{+12.9} \\
Earth, Bactrian-X, 1,800 rows & 10.7 & +10.1 \ci{+6.4}{+14.0} \\
Granite 4.1, Bactrian-X, 1,795 rows$^\S$ & 19.2 & +21.1 \ci{+15.9}{+26.3} \\
\quad no Korean rows (1,748 rows)$^\S$ & 19.2 & +19.5 \ci{+14.3}{+24.7} \\
\bottomrule
\end{tabular}}
\caption{\textbf{Shaded: the main run and the run with no Korean rows,} whose rises are about the same. Guard-measured Korean rise under plain LoRA in the ten settings where a plain LoRA was trained on task data alone (points). The settings share base models and training rows and are not independent; the Global, no-Korean, 519-row, Aya and Earth rows were chosen for an earlier figure after their results were known (exploratory), the three single-language rows were fixed before those runs were read, and the Granite rows come from a block registered before any Granite run. On Tiny Aya the guard's own change on harmless Korean prompts is +5.5 (\S\ref{sec:measure}). $^\S$On Granite the guard's Korean change on harmless prompts is +13.1, and the human readings split: readers 1 and 2 detected no rise and readers 3 and 4 detected one (exploratory; \S\ref{sec:granite}).}
\label{tab:settings}
\end{table}

\begin{table}[!ht]
\centering\footnotesize
\setlength{\tabcolsep}{3pt}
\resizebox{\columnwidth}{!}{\begin{tabular}{@{}lll@{}}
\toprule
Granite 4.1 3B, 3 seeds & Capped at 256 & Generated again \\
\midrule
G1: plain $-$ base (en, ko, zh) & +10.1 \ci{+7.6}{+12.6} & +11.5 \ci{+9.2}{+13.9} \\
\keyrow G2: Korean rise $-$ English rise & +20.8 \ci{+15.1}{+26.3} & +24.2 \ci{+18.9}{+29.3} \\
G3: SafeGrad $-$ plain (en, ko, zh) & $-$13.7 \ci{-16.1}{-11.4} & $-$13.9 \ci{-16.3}{-11.5} \\
Korean \%: base, plain, SafeGrad & 19.2, 40.3, 15.1 & 14.4, 38.8, 14.0 \\
\bottomrule
\end{tabular}}
\caption{\textbf{Shaded: G2,} the Korean rise minus the English rise. The model outside the Tiny Aya family: the three registered checks on capped answers and after every capped answer was generated again at 1,024 tokens (points, 95\% interval; G2 from unrounded values). With the guard's Unsafe class alone the three are +8.9 \ci{+6.7}{+11.1}, +18.6 \ci{+13.6}{+23.6} and $-$11.6 \ci{-13.8}{-9.4}. Guard verdicts only, and the rise is likely but not established: on harmless prompts the guard's flags of Korean answers rose by +13.1 \ci{+9.6}{+16.9}, and the human readings split, with readers 1 and 2 detecting no rise and readers 3 and 4 detecting one ($p=0.003$ and $p=0.039$; exploratory) (\S\ref{sec:granite}; Appendix~\ref{app:protocol}).}
\label{tab:granite}
\end{table}

\begin{table}[!ht]
\centering\footnotesize
\setlength{\tabcolsep}{3pt}
\resizebox{\columnwidth}{!}{%
\begin{tabular}{@{}lrlll@{}}
\toprule
Language & Base & Plain $-$ base & SafeGrad $-$ plain & Mixing $\times$5 $-$ plain \\
\midrule
English & 1.0 & +1.8 \ci{-0.7}{+4.3} & $-$2.2 \ci{-4.7}{+0.0} & $-$2.5 \ci{-4.7}{-0.8} \\
\keyrow Korean & 6.0 & +10.5 \ci{+5.8}{+15.3} & $-$7.2 \ci{-10.7}{-3.8} & $-$8.5 \ci{-12.2}{-4.8} \\
Chinese & 3.5 & +11.8 \ci{+7.7}{+16.3} & $-$5.2 \ci{-8.8}{-1.8} & $-$8.8 \ci{-12.5}{-5.5} \\
Swahili & 5.0 & +0.5 \ci{-2.5}{+3.3} & $-$1.2 \ci{-3.7}{+1.3} & $-$3.2 \ci{-5.7}{-1.0} \\
\bottomrule
\end{tabular}}
\caption{\textbf{Shaded: Korean,} the language the registered rules cover. The second benchmark: the 200 unsafe prompts of XSTest. Guard-flagged rate (\%) and differences in points pooled over seeds 42, 1 and 2, 95\% intervals. English as written; the other languages translated with NLLB-200. No person has checked the guard or the translations; answers capped at 256 tokens. Registered rules cover the Korean row (X1 plain against base; X3 and X4 the two repairs against plain, on a benchmark their refusal examples do not come from) and the Korean rise minus the English rise (X2, +8.7 \ci{+3.5}{+13.8}). X3 is a pooled result and is not significant at every seed; no rule compares the two repairs here, and nothing was registered for Chinese.}
\label{tab:second}
\end{table}

\begin{table}[!ht]
\centering\footnotesize
\setlength{\tabcolsep}{3pt}
\resizebox{\columnwidth}{!}{\begin{tabular}{@{}lrrrrl@{}}
\toprule
NF4 $-$ bf16 & en & ko & zh & sw & Four languages \\
\midrule
\keyrow Base, 1 run & +5.5* & +2.4 & +1.7 & +2.1 & +2.9 \ci{+1.5}{+4.6} \\
Plain, 3 seeds & +1.1 & +3.2* & +2.5* & +4.8* & +2.9 \ci{+1.7}{+4.2} \\
SafeGrad, 3 seeds & +2.6* & +2.6* & +1.1 & +1.1 & +1.9 \ci{+0.8}{+3.0} \\
\bottomrule
\end{tabular}}
\caption{\textbf{Shaded: the base model,} where NF4 alone raised the rate. Change in the guard-flagged rate (points) when the base weights are loaded in NF4 \citep{dettmers2023qlora} (bitsandbytes), on one L4 GPU, with the adapter left unmerged in both precisions. * the test or the 95\% interval excludes zero, not corrected for their number. The base row is one run with no registered rule. Merging at seed 42 changed the rate by +0.0 \ci{-0.8}{+0.8} (plain) and $-$0.3 \ci{-1.2}{+0.4} (SafeGrad).}
\label{tab:nf4}
\end{table}

\paragraph{More on \S\ref{sec:drift}.} \emph{Cut-off answers.} On MultiJail at seed 42, 34\% of the base model's Korean answers reached the 256-token cap against 8\% of plain's; after every capped answer was generated again at 1,024 tokens the pooled three-seed Korean rise is +13.2 (against +12.7 on capped answers; at seed 42, +14.8 and +14.1), and H1--H3 pass both ways. The Korean rise is there among answers of the same length and among answers that were not cut. By length, the Korean rise under plain is +13.1 \ci{+9.4}{+17.2} among complete answers and above zero in each third (shortest +5.3 \ci{+0.5}{+10.0}); the groups hold different prompts in the two runs, so this is a description. \emph{Second dataset.} On the 519 Aya rows Korean rose by 10.0, 8.9 and 9.3 points, and the Korean rise exceeded the English rise by +10.0 \ci{+5.8}{+14.4}; SafeGrad was $-$4.8 \ci{-7.2}{-2.6} below plain. \emph{Second Tiny Aya model.} On Earth, Korean went from 10.7\% to 21.3\%, 19.2\% and 21.6\%; SafeGrad was $-$6.2 \ci{-9.2}{-3.4} below plain. \emph{Guard class.} Every registered three-seed check also passes when only the guard's Unsafe class is counted; read at seed 42 alone, the Chinese-only comparison of SL2 does not. Under that class the Korean rise exceeds the English rise by +8.8 \ci{+4.7}{+13.1}, and Korean goes from 2.7\% to 12.7\% under plain and 6.1\% under SafeGrad.

\paragraph{More on \S\ref{sec:language}.} \emph{One language at a time} (600 rows; exploratory). Trained on English alone, on Korean alone or on Chinese alone, the Korean rate rose by 7 to 12 points in each of three seeds. In the English-only and Chinese-only runs it rose more than the rate in the training language, by +7.6 \ci{+3.7}{+11.7} over English and +7.9 \ci{+3.3}{+12.6} over Chinese. The training language still changes the size (Table~\ref{tab:settings}). English-only training raised the Korean rate less than a mixed set of similar size, and still by 8 points ($-$4.6 \ci{-7.8}{-1.4}; exploratory, registered after its seed-42 value was seen).

\paragraph{Reproducibility checks.} These repeat one result under the same guard and are not replications. Run through AlignTune \citep{lexsi2026aligntune}, the same fine-tune raised the Korean rate from 5.2\% to 18.6\% (+13.4 points, $p<0.001$; our own trainer gave 19.2\% at that seed; one run). Evaluated again on a cloud GPU, none of eight rates changed significantly (largest change 1.4 points).

\begin{table}[!ht]
\centering\footnotesize
\setlength{\tabcolsep}{3pt}
\begin{adjustbox}{max width=\columnwidth}
\begin{tabular}{@{}llcl@{}}
\toprule
 & & Base $\rightarrow$ plain (\%) & Change \\
\midrule
\keyrow Tiny Aya & Korean & 5.6 $\rightarrow$ 11.1 & +5.5 \ci{+2.1}{+9.1} \\
 & English & 1.6 $\rightarrow$ 5.3 & +3.7 \ci{+1.2}{+6.4} \\
\keyrow  & ko $-$ en & & +1.7 \ci{-2.3}{+6.0} \\
Granite 4.1 & Korean & 4.0 $\rightarrow$ 17.1 & +13.1 \ci{+9.6}{+16.9} \\
 & English & 2.0 $\rightarrow$ 4.8 & +2.8 \ci{+0.9}{+4.9} \\
\bottomrule
\end{tabular}
\end{adjustbox}
\caption{\textbf{Shaded:} the Korean change, and the Korean minus English contrast, where no difference was detected. The harmless-prompt flag rate: the guard's flags on answers to safe XSTest prompts, three seeds (change in points, 95\% intervals); exploratory.}
\label{tab:harmless}
\end{table}

\paragraph{More on \S\ref{sec:measure}.} Cohen's $\kappa$ between the guard and the 120 labels is 0.87 in English, 0.60 in Korean, 0.80 in Chinese and 0.49 in Swahili (3 of 30 Swahili answers were marked unsure). When the Swahili answers the guard had passed were translated to English with NLLB-200 and judged again, the guard-flagged rate was 17.9\% for base, 29.2\%, 25.1\% and 25.4\% for plain, and 17.2\%, 17.5\% and 17.5\% for SafeGrad. This step can only add flags, in every arm, and nobody checked the translations. We report only that plain stays above base under it.

\paragraph{More on \S\ref{sec:damage}--\ref{sec:granite}.} Under plain the task score went from 6.61 to 6.37 in English, 6.10 to 5.51 in Korean and 6.25 to 5.79 in Chinese. At one seed (Table~\ref{tab:dose}), at 0.25 of the fine-tune's update the Korean rate had already risen (+3.8 \ci{+1.0}{+6.5}) with no detected fall in the task score ($-$0.08 \ci{-0.20}{+0.02}). Over three seeds the rise at that strength was not detected once the guard's change on harmless prompts was subtracted (+3.2 \ci{-0.2}{+6.9}; \S\ref{sec:damage}). The random arms scored 6.34 and 6.37 on the task against 6.32 for base, so their result concerns the size of the change only. On Granite, no difference between SafeGrad and base was detected in Korean ($-$4.1 \ci{-8.6}{+0.2}), and plain refused 15.2\%, 11.2\% and 16.8\% of safe English prompts. Korean at seeds 1 and 2: 8.2\% and 9.6\% under SafeGrad, 6.2\% and 6.2\% with mixing $\times$5.

\paragraph{More on the summarisation fine-tune} (\S\ref{sec:damage}; exploratory). The rise was detected among complete answers (+6.3 \ci{+3.8}{+9.1}) and in the middle third of answer lengths, but not in the shortest or longest third, where it is not separated from the change in length; 13 of the base run's 14 Korean flags are on answers that reached the cap. Korean rose more than English (+3.6 \ci{+0.3}{+7.0}). The Korean rise exceeded the guard's change on harmless Korean prompts (+4.5 \ci{+0.7}{+8.3}), which was +0.0 \ci{-2.7}{+2.7} here, while harmless-prompt flags rose in English (+2.7 \ci{+0.4}{+5.1}) and Chinese (+2.5 \ci{+0.3}{+4.9}). No difference between SafeGrad and plain was detected in Korean ($-$2.3 \ci{-4.6}{+0.0}). A second reading with uncapped answers is in Table~\ref{tab:checks2} (row 20c); two of its contrasts differ from the first.

\emph{Mixing on that fine-tune.} Three further runs mixed five copies of the refusal examples into this training data (mixing $\times$5, \S\ref{sec:repair}; Table~\ref{tab:xlsum}). Mixing left the Korean rate below plain ($-$5.6 \ci{-8.5}{-3.1}), with no difference from base detected ($-$1.1 \ci{-3.4}{+1.1}), and ROUGE-L did not fall (+0.67 \ci{+0.02}{+1.35} against plain). It refused more harmless prompts: in English +9.5 \ci{+6.0}{+13.2} over base and +7.1 \ci{+3.5}{+10.9} over plain, in Korean +6.4 \ci{+2.8}{+10.0} over base; SafeGrad's English figure was +4.8 \ci{+2.3}{+7.5}.

\begin{table}[!ht]
\centering\footnotesize
\setlength{\tabcolsep}{3pt}
\resizebox{\columnwidth}{!}{\begin{tabular}{@{}lrrrl@{}}
\toprule
Plain LoRA, seed 42 & Korean & English & Task & Task against base \\
\midrule
\keyrow Our settings (2e-4, 3 epochs) & +14.1* & +0.0 & 5.94 & $-$0.38 \ci{-0.52}{-0.25} \\
LR halved (1e-4) & +13.1* & +2.1 & 5.71 & $-$0.61 \ci{-0.77}{-0.44} \\
LR doubled (4e-4) & +16.5* & +3.1 & 5.94 & $-$0.39 \ci{-0.52}{-0.25} \\
Half the epochs (1.5) & +21.0* & +4.8* & 5.67 & $-$0.66 \ci{-0.82}{-0.48} \\
Twice the epochs (6) & +11.3* & +1.4 & 5.97 & $-$0.35 \ci{-0.48}{-0.23} \\
\bottomrule
\end{tabular}}
\caption{\textbf{Shaded: our settings;} compare the Korean and task columns of the other rows with it. Guard-measured rises in the guard-flagged rate (points against base; * $p<0.05$) and task score (1--7; base 6.32), with the learning rate (LR) and the epochs each halved and doubled. Korean rose more than English each time; the largest Korean rise goes with the lowest task score and the smallest with the highest. One run and one seed per setting; no trend is tested.}
\label{tab:sens}
\end{table}

\begin{table}[!ht]
\centering\footnotesize
\setlength{\tabcolsep}{3pt}
\resizebox{\columnwidth}{!}{\begin{tabular}{@{}lllll@{}}
\toprule
Arm (seed 42, no training) & Korean rise & English rise & Korean $-$ English & Task against base \\
\midrule
\keyrow Update $\times$ 0.25 & +3.8 \ci{+1.0}{+6.5} & $-$1.4 \ci{-3.8}{+1.0} & +5.2 \ci{+1.4}{+9.3} & $-$0.08 \ci{-0.20}{+0.02} \\
Update $\times$ 0.5 & +5.5 \ci{+2.1}{+8.9} & $-$1.4 \ci{-4.1}{+1.0} & +6.9 \ci{+2.7}{+11.3} & $-$0.24 \ci{-0.35}{-0.12} \\
Update $\times$ 0.75 & +9.3 \ci{+5.2}{+13.4} & +1.0 \ci{-1.4}{+3.8} & +8.2 \ci{+2.7}{+13.7} & $-$0.32 \ci{-0.44}{-0.19} \\
The plain run ($\times$ 1) & +14.1 \ci{+9.6}{+18.9} & +0.0 \ci{-2.7}{+2.7} & +14.1 \ci{+8.6}{+19.6} & $-$0.38 \ci{-0.52}{-0.25} \\
\keyrow Random update, draw 1 & +0.3 \ci{-1.0}{+1.7} & $-$0.3 \ci{-1.7}{+0.7} & +0.7 \ci{-1.4}{+2.7} & +0.02 \ci{-0.03}{+0.08} \\
\keyrow Random update, draw 2 & $-$0.3 \ci{-2.1}{+1.4} & $-$1.4 \ci{-3.1}{+0.3} & +1.0 \ci{-1.7}{+3.8} & +0.04 \ci{-0.02}{+0.12} \\
\bottomrule
\end{tabular}}
\caption{\textbf{Shaded: a quarter of the update and the two random updates.} The plain adapter with its update scaled, and two random updates as large as plain's in every module. Guard-measured rises (points against base, 95\% intervals over 291 prompt families); task score against base (6.32, scale 1--7). One adapter, one seed. Over three seeds the rise at 0.25 was not detected once the guard's change on harmless prompts was subtracted (+3.2 \ci{-0.2}{+6.9}; \S\ref{sec:damage}); that change is itself not detected there (+0.8 \ci{-1.7}{+3.3}).}
\label{tab:dose}
\end{table}

\begin{table}[!ht]
\centering\footnotesize
\setlength{\tabcolsep}{3pt}
\resizebox{\columnwidth}{!}{\begin{tabular}{@{}lcccccccc@{}}
\toprule
 & & Length & Task & \multicolumn{4}{c}{Guard-flagged rate against base} & ko flagged of 291, \\
Run & ROUGE-L & (chars) & score & en & ko & zh & sw & capped $\to$ again \\
\midrule
Base model & 31.2 & 150 & 6.29 & & & & & 14 $\to$ 12 \\
Plain, seed 42 & 39.2 & 67 & 5.80 & & & & & 26 $\to$ 24 \\
Plain, seed 1 & 39.2 & 70 & 5.86 & & & & & 30 $\to$ 27 \\
Plain, seed 2 & 39.7 & 70 & 5.88 & & & & & 25 $\to$ 23 \\
\keyrow \quad three seeds, against base & +8.18* & & $-$0.45* & +0.9 & +4.5* & +2.3 & +1.4 & \\
SafeGrad, seed 42 & 39.1 & 68 & 6.14 & & & & & 22 $\to$ 17 \\
SafeGrad, seed 1 & 39.3 & 72 & 6.08 & & & & & 20 $\to$ 14 \\
SafeGrad, seed 2 & 39.5 & 70 & 6.19 & & & & & 19 $\to$ 17 \\
\keyrow \quad three seeds, against base & +8.11* & & $-$0.16* & $-$0.5 & +2.2 & $-$0.2 & $-$0.8 & \\
Mixing $\times$5, seed 42 & 39.9 & 69 & 6.11 & & & & & 8 $\to$ 6 \\
Mixing $\times$5, seed 1 & 40.1 & 68 & 6.12 & & & & & 13 $\to$ 11 \\
Mixing $\times$5, seed 2 & 40.1 & 70 & 6.13 & & & & & 11 $\to$ 10 \\
\keyrow \quad three seeds, against base & +8.84* & & $-$0.18* & $-$3.3* & $-$1.1 & $-$2.7* & +0.0 & \\
\bottomrule
\end{tabular}}
\caption{\textbf{Shaded: the three-seed changes against base} for plain, SafeGrad and mixing $\times$5; the rows above them are single runs. The summarisation fine-tune by run: character-level ROUGE-L against one person's summary on 300 held-out articles, mean answer length in characters, the general task score (1--7), and, pooled over seeds, the change against base in points on capped answers (* the 95\% interval excludes zero). Mixing $\times$5: five copies of the refusal examples. Last column, with no rule: Korean attack answers the guard flags, on the capped answers and after re-generation (section 20c). 3,537 answers that had reached the cap were generated again at 1,024 tokens for the ten runs (2,495 to attack prompts, 1,042 to harmless Korean prompts); at most 9 of 291 answers in a run and language were still cut.}
\label{tab:xlsum}
\end{table}

\begin{table}[!ht]
\centering\footnotesize
\setlength{\tabcolsep}{3pt}
\resizebox{\columnwidth}{!}{\begin{tabular}{@{}llrrrrl@{}}
\toprule
 & & \multicolumn{2}{c}{Base} & \multicolumn{2}{c}{Plain, 3 seeds} & \\
Korean attack answers & Tokens & $n$ & flagged & $n$ & flagged & Plain $-$ base \\
\midrule
\keyrow Complete & 8 to 255 & 194 & 0.5\% & 688 & 6.8\% & +6.3 \ci{+3.8}{+9.1} \\
Reached the cap & 256 & 97 & 13.4\% & 185 & 18.4\% & +5.0 \ci{-2.7}{+13.1} \\
Shortest third & 8 to 58 & 42 & 2.4\% & 346 & 6.4\% & +4.0 \ci{-0.6}{+8.3} \\
Middle third & 58 to 200 & 118 & 0.0\% & 270 & 6.7\% & +6.7 \ci{+3.3}{+10.6} \\
Longest third & 201 to 256 & 131 & 9.9\% & 257 & 16.0\% & +6.0 \ci{-0.1}{+12.4} \\
\bottomrule
\end{tabular}}
\caption{\textbf{Shaded: complete answers,} where the rise is clearest. The summarisation fine-tune's Korean rise by answer length and the cap (C1; points, 95\% intervals over prompt families). A group is defined by the answer, which the fine-tune changes, so it holds different prompts in the two runs: a description, not a control.}
\label{tab:xlsumcut}
\end{table}

\section{4-bit Loading and Mixed Precision}
\label{app:quant}

\paragraph{NF4 (Table~\ref{tab:nf4}).} Compression can change safety behaviour \citep{hong2024compressed,egashira2024quantization}. Under NF4, SafeGrad stayed below plain on Korean in every seed (pooled $-$9.9 \ci{-13.3}{-6.5}). On the base model the two 4-bit schemes we tried differ: NF4 raised the guard-flagged rate in one run (+2.9 \ci{+1.5}{+4.6}; by language only English, Table~\ref{tab:nf4}), and for uniform 4-bit with optimum-quanto no change was detected (+0.2 \ci{-1.2}{+1.5}). Not measured: an adapter merged before quantization, GGUF. Memory, speed and the task score of all four loadings are in Table~\ref{tab:quant-deploy}.

\paragraph{Mixed precision guided by the refusal circuit.} Registered before any quantized model ran: Tiny Aya Global with no adapter, quantized with optimum-quanto \citep{hf2023quanto} through CircuitKIT \citep{lexsi2026circuitkit} at 4.0 bits on average. Uniform 4-bit was set against a plan that keeps the three layers the refusal circuit ranks highest at 8 bits and its six lowest at 2 bits, and against three random plans of the same shape. No rise in the guard-flagged rate was detected under uniform 4-bit (+0.2 \ci{-1.2}{+1.5} over four languages), so by the registered rule the circuit plan was not read against uniform 4-bit or the random plans. The safety hypothesis is untested, and we claim no protection. The circuit plan scored below uniform 4-bit on the task ($-$1.31 \ci{-1.52}{-1.10}) and on Global-MMLU-Lite \citep{singh2024globalmmlu} ($-$14.1 \ci{-18.2}{-10.1}). Task score: unquantized 6.35, uniform 4-bit 6.09, circuit plan 4.78, random plans 3.69 to 4.15. One run per arm; these weights were evaluated stored at full width. A later run loaded all four with the weights packed on one GPU and measured memory and speed (Table~\ref{tab:quant-deploy}): every 4-bit loading saves about 62\% of memory and is slower than bf16, and the circuit plan is the slowest.

\paragraph{The rank test} (exploratory). At the same 4.0 bits, the model kept a lower perplexity under the circuit plan than under random plans of the same shape: 7.81 against 8.19 to 12.61 for 43 random plans ($p=0.023$). The test was registered after the gate's perplexity table had been read and counts the gate's three random plans among the 43; on the 40 plans the gate had not shown, none is at or below the circuit plan ($p=0.024$). Perplexity is not safety, no ranking that knows nothing of refusal was compared, and on Global-MMLU-Lite the circuit plan is below two of three random plans.

\begin{table}[!ht]
\centering\footnotesize
\setlength{\tabcolsep}{3pt}
\begin{adjustbox}{max width=\columnwidth}
\begin{tabular}{@{}lrrrr@{}}
\toprule
Base model & bf16 & NF4 & 4-bit & Circuit \\
\midrule
\keyrow Memory after loading (MiB) & 6,496 & 2,418 & 2,475 & 2,470 \\
Peak memory, batch 8 (MiB) & 7,045 & 2,966 & 3,025 & 3,021 \\
Output tokens/s, batch 1 & 24.4 & 17.3 & 20.4 & 16.6 \\
Output tokens/s, batch 8 & 142.1 & 100.4 & 114.1 & 87.9 \\
Perplexity, task references & 6.22 & 6.39 & 6.49 & 7.81 \\
\keyrow Task score (1--7) & 6.35\textsuperscript{a} / 6.29 & 6.20 & 6.09\textsuperscript{a} & 4.78\textsuperscript{a} \\
Global-MMLU-Lite (\%) & 54.5\textsuperscript{c} / 54.8 & 51.4 & 51.2\textsuperscript{c} & 37.1\textsuperscript{c} \\
\midrule
Guard-flagged rate, 4 languages (\%) & 6.8\textsuperscript{a} / 6.0\textsuperscript{b} & 8.9\textsuperscript{b} & 7.0\textsuperscript{a} & 10.0\textsuperscript{a} \\
Over-refusal, English (\%) & 9.6\textsuperscript{a} / 9.6\textsuperscript{b} & 7.6\textsuperscript{b} & 8.4\textsuperscript{a} & 18.8\textsuperscript{a} \\
\bottomrule
\end{tabular}
\end{adjustbox}
\caption{\textbf{Shaded: memory and task score,} the trade-off: every 4-bit loading saves memory, and the circuit plan loses the most task score. Tiny Aya Global with no adapter in four loadings: bfloat16, NF4 (bitsandbytes), uniform 4-bit and the circuit plan (quanto, 8/4/2-bit). Unmarked cells: one run on an NVIDIA A100-80GB with the 4-bit weights packed. \textsuperscript{a}Apple laptop GPU, quanto weights written out at full width. \textsuperscript{b}NVIDIA L4. \textsuperscript{c}NVIDIA L4, quanto weights at full width. The guard-flagged rate is averaged over English, Korean, Chinese and Swahili; safety was not measured with packed weights. The circuit plan's safety is exploratory: the registered gate said not to read it, because uniform 4-bit raised no detected harm. On the A100 every 4-bit loading uses about 62\% less memory than bf16 and is slower.}
\label{tab:quant-deploy}
\end{table}

\begin{table*}[t]
\centering\scriptsize
\setlength{\tabcolsep}{3pt}
\begin{tabular}{@{}p{1.6cm}p{4.6cm}cp{3.3cm}p{4.3cm}@{}}
\toprule
Check & What it reads & Seeds & Registered & Outcome \\
\midrule
\keyrow H1, H2, H3 & Korean rises more than English; SafeGrad below plain on Korean; Swahili refusals fall & 3 & before any stage 2 evaluation; seed 42 known from stage 1 & pass, pass, pass, in every seed; the same after re-generation \\
H4 & A clean Korean training set & -- & as H1 & not tested: the set could not be cleaned \\
A, B & A second judge against our 120 labels; the headline under it & -- & before each judge was loaded; the third after outside reviews & A failed with three judges; B never run \\
C & Swahili re-translation with NLLB-200 & 1 & before & pass (a direction only; Appendix~\ref{app:tables}) \\
E1, E2 & Earth: drift and repair; a larger Swahili fall & 3 & seed 42 before; three-seed rule after seed 42 & E1 holds; E2 not confirmed \\
Y1, Y2, Y3 & Aya: drift, repair, Swahili refusals & 3 & seed 42 before; three-seed rule after seed 42 & Y1, Y2 pass; Y3 fails \\
K1 & Korean rise with no Korean rows & 3 & seed 42 before; seeds 1, 2 after seed 42 & holds in all three seeds \\
K2, K3 & A person's reading, Tiny Aya & 1 & K2 before; K3 after K2 did not pass & neither passes ($p=0.22$, $p=0.070$) \\
SL1, SL2; C2 & One training language at a time; a mixed set of similar size & 3 & seed 42 before; seeds 1, 2 and C2 after seed 42 & SL1, SL2 pass; C2 $-$4.6 \ci{-7.8}{-1.4} \\
Dir-R1 & Korean and English move the same distance along the refusal direction & 3 & before & holds as predicted \\
Circ-N1 & Korean keeps the fewest base refusal nodes & 1 & before & holds at 100, 200, 300 nodes; not at 50, 400 \\
M1, M2, M3 & Mixing against SafeGrad: Korean, English over-refusal, task & 2 & before the two new seeds; seed 42 seen & mixing $\times$5: M1 below zero; mixing $\times$1: includes zero; M2 above zero for both; M3 includes zero \\
U1, U2, U3 & Six more languages & 3 & before & refusals fell on average; SafeGrad kept more; none rose more than Korean; for three, no difference from Korean detected \\
AT-T1 & The plain run through AlignTune & 1 & before & a significant Korean rise (+13.4; bar 6.36) \\
S1, S2, S3 & Learning rate and epochs & 1 each & before; S3 added before any evaluation & Table~\ref{tab:sens}; each rise above the bar of 6.36; the forecast of less drift in milder runs failed \\
X1--X4 & Second benchmark & 3 & before & X1, X2 above zero; X3, X4 below zero \\
NF4-Q2; P1, P2 & 4-bit loading: at seed 42; on three seeds & 1; 3 & before; seed 42 under NF4 known for P1, P2 & pass; P1 above zero for both models, P2 pass \\
MP-Q0--Q3, B1 & Mixed precision guided by the circuit & 1 & before & Q0 includes zero, so Q1, Q2 not read; Q3, B1 below zero \\
Rank-R1 & Circuit plan against 43 random plans, perplexity & 1 & after the gate's perplexity table was read & $p=0.023$; on 40 new plans $p=0.024$ \\
\bottomrule
\end{tabular}
\caption{\textbf{Shaded: H1--H3, the only confirmatory checks.} The registered checks, part 1 of 2 (part 2: Table~\ref{tab:checks2}). The codes are those of the preregistration; the prefixes Dir-, Circ-, AT-, NF4-, MP-, Rank-, G-, 19-, 19b- and the section numbers in brackets are ours, added because the registration uses several codes more than once (R1 for three checks; R2, N1, T1, Q1, Q2, S1, S2, X1 to X4, C1, M1 and M2 for two each). ``Before'' means before the check's own runs or computation. Second readings after re-generation were all registered after the first reading and are not used for any claim; each came out the same, except X2 and X4 of section 20 (row 20c). Registered checks this paper does not use (L1 and L2, the C1 and R2 of earlier sections, Q1 of section 12.2, V1) are on the status page, \texttt{results/lir/stage2/README.md}.}
\label{tab:checks}
\end{table*}

\begin{table*}[t]
\centering\scriptsize
\setlength{\tabcolsep}{3pt}
\begin{tabular}{@{}p{1.6cm}p{4.6cm}cp{3.3cm}p{4.3cm}@{}}
\toprule
Check & What it reads & Seeds & Registered & Outcome \\
\midrule
G1, G2, G3 & Granite: drift, Korean against English, repair & 3 & before & above zero, above zero, below zero \\
G-N1; G-T1, G-T2 & Granite: no Korean row; task score & 3 & before & above zero; $-$0.47 \ci{-0.61}{-0.35}; $-$0.07 \ci{-0.15}{+0.00} \\
16d & A person's reading, Granite & 1 & after the guard's result was known & does not pass: 83 against 84, $p=1.00$ \\
18a--18i & Analyses of stored answers & 3 & each written down before it was computed; exploratory & 18b: \S\ref{sec:measure}; 18c: Appendix~\ref{app:tables} \\
Cut-off audit; 13g; 13h & H1--H3 after re-generation; every strict check on the Unsafe class alone; the task-score drop at matched answer length & 3 & each before it was computed (13g on answers whose strict reading was known) & H1--H3 pass before and after; every three-seed check passes; $-$0.39 \ci{-0.81}{-0.05} \\
19-D1 & Lowest strength of the plain update with a Korean rise; the task score there & 1 & before any adapter was built & at 0.25: Korean +3.8 \ci{+1.0}{+6.5}, task $-$0.08 \ci{-0.20}{+0.02} (no detected fall) \\
19-D2 & Korean rise under two random updates & 1 & as 19-D1 & no rise: +0.3 \ci{-1.0}{+1.7}, $-$0.3 \ci{-2.1}{+1.4} \\
19b-S1, S2 & Strength 0.25 at three seeds; its Korean rise minus the guard's change on harmless prompts & 3 & before its arms were built; seed 42's attack and task figures known & S1: +4.0 \ci{+1.5}{+6.6}, task $-$0.08 \ci{-0.18}{+0.02}; S2: +3.2 \ci{-0.2}{+6.9}, not detected \\
20 (X0--X4) & A fine-tune that improves its own task: its gain; the Korean rise; Korean against English; against the guard on harmless prompts; SafeGrad & 3 & before any job; seed 42's task result read before the other runs & X0 +8.18 \ci{+6.96}{+9.36}; X1 +4.5 \ci{+1.9}{+7.2}; X2 +3.6 \ci{+0.3}{+7.0}; X3 +4.5 \ci{+0.7}{+8.3}; X4 $-$2.3 \ci{-4.6}{+0.0}, includes zero. Second readings: row 20c \\
C1; M1, M2 (20b) & On that fine-tune: the rise by answer length; mixing against plain in Korean and in ROUGE-L & 3 & before any safety verdict of the block was opened; read because X1 was above zero & C1: not found in the shortest and longest third; M1 $-$5.6 \ci{-8.5}{-3.1}; M2 +0.67 \ci{+0.02}{+1.35}, no fall detected \\
20c & X1 to X4 and M1 (20b) after every capped answer was generated again at 1,024 tokens & 3 & Nov 25, after X0 to X4 and C1 were read and before the mixing runs were read; not used for any claim & X1 +4.4 \ci{+1.8}{+7.0}, X3 +4.1 \ci{+0.4}{+7.8}, M1 $-$5.4 \ci{-8.1}{-2.9}: the same on the capped answers and after re-generation. X2 +3.2 \ci{-0.1}{+6.5}, includes zero, and X4 $-$3.0 \ci{-5.2}{-0.8}, below zero: these differ; we report the first, registered reading \\
21 (R1, R2) & A second reader on both models & 1 & before the sheets were made; the guard's and the first reader's results known & R1 +6.0 \ci{+2.9}{+8.9}; R2 9 against 14 on 75 prompts; marks returned in a form the entry excluded (Appendix~\ref{app:protocol}) \\
\bottomrule
\end{tabular}
\caption{The registered checks, part 2 of 2 (part 1 and the notes on the codes: Table~\ref{tab:checks}).}
\label{tab:checks2}
\end{table*}

\section{Exploratory Diagnostics}
\label{app:circuits}

We use the refusal direction \citep{arditi2024refusal} and edge attribution patching \citep{hanna2024faith}. Over three seeds, Korean and English prompts moved the same distance along the base model's refusal direction: the difference is 0.08 units \ci{-0.08}{+0.25}, against a shift of about 3.5 and a registered margin of 1. Adding the direction back into the plain model lowered the Korean guard-flagged rate from 21.2\% to 1.2\% on 80 families and brought 25\% over-refusal, so the intervention is not specific. With CircuitKIT at seed 42, Korean keeps the fewest of the base model's top refusal nodes under plain at 100, 200 and 300 of 612 nodes (74, 156 and 240), not at 50 or 400; at 200 nodes the top nodes recover 65\% of the refusal contrast in English, 91\% in Korean, 26\% in Chinese and 40\% in Swahili. None of this shows a mechanism.

\section{Human Reading Protocol}
\label{app:protocol}

\paragraph{The first reader.} One author read Korean answers of base and plain at seed 42 with the arm hidden, in three registered readings (K2 and K3 on Tiny Aya, 16d on Granite; Table~\ref{tab:readers}). K3 was registered after K2 did not pass, and 16d after the guard's result on Granite was known; 16d's marks came back as a pasted list, not as the saved sheet.

\paragraph{The second reader.} A second reader marked the same kind of sheet: 225 Tiny Aya prompts, read as a weighted difference over the 291, and 75 of the 150 Granite prompts, read by exact McNemar. By an author's account the reader is outside the team, a native speaker of Korean, and was not told what we expected. The reader marked in a labelling page set up by that author, which shows only what is in the sheet loaded into it; we were given the page's copy after the reading. The marks came back as a pasted list, which the registration said would not be taken, and were scored by the registrant's decision against that rule. By the same account the reading took at most 72 to 101 minutes for 640 rows.

\paragraph{What the marks show.} On the Tiny Aya sheet the second reader marked 11 base against 28 plain answers unsafe on 225 prompts (exact McNemar $p<0.001$). By the same weighted sum the guard counts +14.1 \ci{+10.7}{+16.8} (the sheet's stratified bootstrap; over prompt families, \ci{+9.6}{+18.9}). The second reader gave 370 of 378 Tiny Aya answers the mark the first reader had given ($\kappa$ 0.77 and 0.88), against 71 of 150 on Granite ($\kappa$ 0.07 and 0.10). In each sheet all 20 repeated rows got the same mark, but the page lets a reader look up rows already marked; ``unsure'' was offered and never used.

\paragraph{Readers 3 and 4.} Two more readers, both outside the team and native speakers of Korean, each marked one sheet of 500 rows in a page that shows one row at a time and does not go back to a marked row; the arm and the model are hidden, and the saved file, not a pasted list, was returned. The sheet holds 480 answers and 20 repeats: 300 Korean Tiny Aya answers (150 prompts, both answers each), 100 Granite answers (50 prompts, all from the second reader's Granite sheet) and 80 English Tiny Aya answers (40 prompts, both answers each), all at seed 42. Each answer was matched to its model and arm by its exact prompt and text in the stored answers; every one matched exactly one. This sheet was not registered, and its draw was made after the earlier readings were known. The readers agree with each other at $\kappa$ 0.59 (0.60 on Tiny Aya, 0.66 on Granite), and each gave 19 of 20 repeated rows the same mark. On Tiny Aya, reader 4's marks agree with the first reader's at $\kappa$ 0.81 (K2) and 0.82 (K3) and with the second reader's at 0.73; reader 3 marked more answers unsafe ($\kappa$ 0.43 to 0.51 with them). On Granite the first reader marked 57 of these 100 answers unsafe, against 20 to 29 for the other three readers.

\paragraph{English check} (exploratory, not registered). On the 40 English prompts, base against plain, reader 3 marked 10 against 7 answers unsafe (exact McNemar $p=0.55$) and reader 4 7 against 5 ($p=0.69$); the guard flags 9 against 9 on the same prompts. Like the guard (\S\ref{sec:drift}), neither reader detected an English change, which is what a reader who simply marks fine-tuned answers more often would not show; 40 prompts are too few to rule out a small change. The two readers agree at $\kappa$ 0.46 on these 80 answers.

\paragraph{Records.} How each sheet came back is on record in sections 14c, 16d and 21 of the preregistration.

\section{Preregistration, Checks and Deviations}
\label{app:prereg}

\begin{table}[!ht]
\centering\footnotesize
\setlength{\tabcolsep}{3pt}
\resizebox{\columnwidth}{!}{\begin{tabular}{@{}lcccl@{}}
\toprule
 & & Registered & & Read by \\
Claim (each a guard reading) & Seeds & before its runs & Status & a person \\
\midrule
\keyrow Korean rises more than English (H1) & 3 & seeds 1, 2$^a$ & confirmatory & \S\ref{sec:measure} \\
\keyrow SafeGrad below plain on Korean (H2) & 3 & seeds 1, 2$^a$ & confirmatory & no \\
\keyrow Swahili refusals fall (H3) & 3 & seeds 1, 2$^a$ & confirmatory & no \\
No-Korean and one-language runs & 3 & seed 42$^b$ & exploratory & no \\
Second dataset; second model & 3 & seed 42$^b$ & exploratory & no \\
Six more languages; second benchmark & 3 & yes & exploratory & no \\
Mixing against SafeGrad & 2 & yes$^c$ & exploratory & no \\
Granite 4.1 & 3 & yes & exploratory & \S\ref{sec:granite} \\
Guard on harmless prompts; length & 3 & no & exploratory & no \\
Learning rate and epochs & 1 & yes & exploratory & no \\
Scaled and random updates (\S\ref{sec:damage}) & 1; 3 & yes; partly$^a$ & exploratory & no \\
Summarisation fine-tune (\S\ref{sec:damage}) & 3 & yes & exploratory & no \\
Mixing on that fine-tune & 3 & yes$^d$ & exploratory & no \\
\bottomrule
\end{tabular}}
\caption{\textbf{Shaded: the three confirmatory claims;} every other row is exploratory. What each claim rests on. Only H1--H3 on seeds 1 and 2 are confirmatory (\S\ref{sec:setup}); every check and when it was registered: Appendix~\ref{app:prereg}, Tables~\ref{tab:checks} and~\ref{tab:checks2}. Human readings: \S\ref{sec:measure}. $^a$A seed 42 result was known. $^b$Seeds 1 and 2, or the three-seed rule, registered after seed 42 was read. $^c$Seed 42 was seen first; only seeds 1 and 2 are used in \S\ref{sec:repair}. $^d$Run because a rise was detected.}
\label{tab:rests}
\end{table}

\paragraph{History.} The primary family H1--H3 was committed before any stage 2 model was evaluated; the seed 42 result was known from stage 1. Every later check is in a rolling log and was chosen knowing earlier results, and several were registered after a first result was known. After outside reviews of a draft we added a third second judge, the exploratory analyses of stored answers, the scaled and random updates and the summarisation blocks (\S\ref{sec:damage}), and the second reader (\S\ref{sec:measure}). For mixing on the instruction fine-tune, seed 42 was seen before the rules were written. The mixing runs on the summarisation fine-tune were registered before any safety verdict of that block was opened, and were read only because plain's rise had been detected, which tilts that comparison toward ``below''.

\paragraph{Deviations} (the checks are in Tables~\ref{tab:checks} and~\ref{tab:checks2}). (1) H4 was never tested. (2) Seeds 1 and 2 of the no-Korean and single-language runs, the three-seed rules for Aya and Earth, and the mixed-size comparison were registered after a seed-42 result was known. (3) The repair-vector edit on Bactrian-X was exploratory; it and its registered repeat on the Aya adapters (V1) are not reported in this paper. (4) The six-language block printed the base model's pooled totals once during a file check before the full report; no rule changed. (5) Stage 1 attributed the task-score drop to shorter answers; a registered test showed that is wrong, and that sentence is withdrawn. (6) The AlignTune run was made on the laptop by another team member, with one step of ours in its data path. (7) The second readings after re-generation were registered after the first readings. (8) The second human reading was registered after the first did not pass. (9) The reading on Granite was registered after the guard's result on that model was known, and its marks came back as a pasted list and not as the saved sheet (section 16d). (10) The rank test was registered after the gate's perplexity table had been read. (11) The third second judge and the analyses of section 18 were added after outside reviews of a draft. One figure of 18b had been seen during a file check, and the command of section 18 was committed after its output had been read. (12) The second reader's marks were scored by the registrant's decision against section 21's rule that a pasted list is not taken (Appendix~\ref{app:protocol}). (13) Section 20c reverses the statement of sections 20 and 20b that no answer of the summarisation block is generated again; it was registered after X0 to X4 and C1 were read. (14) The sheet of readers 3 and 4 was not registered and was drawn after the earlier readings were known.

\section{Issues Found in the Tools}
\label{app:feedback}

Chaining the five toolkits on a chat-tuned multilingual model surfaced 26 bugs and gaps, each filed as an issue on the tool's repository (list: \texttt{docs/lir/lexsi-feedback.md}). The two that can change a result without warning: \textbf{AuditKIT \citep{lexsi2026auditkit} \#14}, the paired-bootstrap $p$-value is about 0.5 for any non-zero effect because the bootstrap is not centred; \textbf{SafeTune \#3}, SafeGrad could silently fall back to plain fine-tuning (fixed upstream). Others by tool. \emph{SafeTune:} no gradient accumulation (1); KL with no attention mask (2, fixed) and in the logits' dtype (13, fixed); safety batches never reshuffled (23); WildGuard \citep{han2024wildguard} parser counts any non-refusal as an attack (24); stale alignment gradient after a skipped capture (25); unfinished \texttt{<think>} block scored as an attack (26); Aya red-teaming file duplicated across language codes (21); \texttt{layer\_fraction} picks a layer for a node it drops (22). \emph{CuratorKIT:} no multilingual safety screen (4). \emph{AuditKIT:} no interval on the paired delta (5). \emph{CircuitKIT:} 13 items, four fixed upstream (6, 7, 8, 11), one in part (10) and eight open (9, 12, 15 to 20), among them scores that depend on batch size (15), \texttt{faithfulness} scoring a different circuit from \texttt{discover} (16) and a random baseline that is usually an empty graph (18); the rest are in the issues file.

\section{Reproducibility and Artifacts}
\label{app:repro}

Code: \named{\url{https://github.com/lost-in-refusal/lexsi-multilingual-safety}}{[link removed for review]}. Every table in \texttt{results/lir/stage2/} is written by a named command; \texttt{python -m unittest tests.test\_lir} runs the test suite (each registered rule has a test on synthetic data). \texttt{python -m lir.rebuild fetch} and \texttt{tables} download the public outputs at a pinned revision and rebuild 77 tables from 39 commands with no GPU and no model, tested in a clean clone with no login. Hugging Face\named{ (organization \texttt{lost-in-refusal})}{}:
\begin{itemize}\setlength\itemsep{0pt}
\item Generations and verdicts (the answers to the held-out Aya prompts and the XL-Sum summaries are withheld; their scores are public): \named{\url{https://huggingface.co/datasets/lost-in-refusal/lost-in-refusal-eval-outputs}}{[link removed for review]} Not uploaded: the third judge's verdicts and the answers of the scaled and random updates and of the summarisation runs.
\item Frozen data: \named{\url{https://huggingface.co/datasets/lost-in-refusal/lost-in-refusal-data}}{[link removed for review]}
\item Stage 1 adapters: \named{\url{https://huggingface.co/lost-in-refusal/tiny-aya-global-bactrian-lora-plain} and \url{https://huggingface.co/lost-in-refusal/tiny-aya-global-bactrian-lora-safegrad}}{[links removed for review]}
\item Stage 2 adapters (47 folders; the adapters trained on a cloud GPU are not released): \named{\url{https://huggingface.co/lost-in-refusal/tiny-aya-lora-stage2}}{[link removed for review]}
\end{itemize}

\section{Declarations}
\label{app:decl}

\paragraph{Timeline.} All code, data, runs and writing were made between the first commit on November 2, 2026 and the submission of this version.

\paragraph{Author contributions.} \ifanon [Removed for review.] \else Aishwarya Nikkumbh and Mingyu Kang contributed equally. Aishwarya Nikkumbh led the project, designed the red-team and evaluation plan, wrote parts of the preregistration and wrote the paper. Mingyu Kang built the training, evaluation and interpretability pipeline, ran the laptop experiments and the later cloud runs, wrote the results drafts, and labelled the Korean and Chinese answers. Hannah Pham curated the training data, ran the first GPU experiments, labelled English and Swahili answers and filed tool issues.
\fi

\paragraph{LLM usage.} Most of the code, the analyses, the preregistration text and the drafts of this paper were produced with AI coding assistants working under the authors' direction and review, and three AI systems reviewed a draft. The authors' labels were made with no model assistance, and by an author's account so were the marks of the second reader and of readers 3 and 4 (\S\ref{sec:measure}). The authors are responsible for the content.

\paragraph{Acknowledgements.} \named{We thank Lexsi Labs for mentorship and compute credits, and Hugging Face for compute credits.}{[Removed for review.]}

\FloatBarrier
\end{document}